\documentclass[10pt,conference,anonymous]{IEEEtran} 
\usepackage{subfig} 
\usepackage{graphicx}
\usepackage{url}
\usepackage{xcolor}
\usepackage{multirow}
\usepackage{listings}% http://ctan.org/pkg/listings
\usepackage{algorithmicx,algpseudocode}
\usepackage{algpseudocode}
\usepackage{array}
\usepackage{enumitem}
\usepackage[linesnumbered,algoruled,boxed,lined]{algorithm2e}
\usepackage{amsmath}
\mathchardef\mhyphen="2D % Define a "math hyphen"
\usepackage{stmaryrd}
\usepackage{cuted}
\usepackage{syntax}
\usepackage{multicol}
\usepackage{authblk}
\begin{document}

\title{Probabilistic Conditional System Invariant Generation with Bayesian Inference} 
%\author{Meriel Stein \\ meriel@virginia.edu
%\and Sebastian Elbaum \\ selbaum@virginia.edu
%\and Lu Feng \\ lu.feng@virginia.edu
%\and Shili Sheng  \\ ss7dr@virginia.edu
%}
%\affil{University of Virginia \\ Charlottesville, Virginia}

\author{Meriel Stein}
\author{Sebastian Elbaum}
\author{Lu Feng}
\author{Shili Sheng}
\affil{ University of Virginia, Charlottesville, Virginia \authorcr {\tt \{meriel, selbaum, lu.feng, ss7dr\}@virginia.edu}\vspace{1.5ex}}

\maketitle

\begin{abstract}
 Invariants are a set of properties over program attributes that are expected to be true during the execution of a program.
Since developing those invariants manually can be costly and challenging,
there are a myriad of approaches that support automated mining of likely invariants from sources such as program traces.
Existing approaches, however, are not equipped to 
capture the rich states that condition the behavior of autonomous mobile robots,
or to manage the uncertainty associated with many variables in these systems.
This means that valuable invariants that appear only under specific states remain uncovered. 
In this work we introduce an approach to infer conditional probabilistic invariants  
to assist in the characterization of the behavior of such rich stateful, stochastic systems.
These probabilistic invariants can encode a 
family of conditional patterns, are
generated using Bayesian inference to leverage observed trace data against 
priors gleaned from previous experience and expert knowledge, and are ranked 
based on their surprise value and information content.
Our studies on two semi-autonomous mobile robotic systems show how the proposed
approach is able to generate valuable and previously 
hidden stateful invariants.  
\end{abstract}


\section{Introduction}
\label{sec:introduction}
%\ms{intro is mostly still applicable. First 2 paragraphs are fine.}

Our community has built a large body of work on likely invariant generation from system traces. 
%or to capture behavioral specifications that track object states at runtime \cite{Ammons:2002} \cite{Ernst2000QuicklyDR} \cite{daikon} or sequential invocation patterns \cite{Yang:2006:PMT}. 
%\se{why just from traces, justify}
%Traces provide an artifact of dynamic/runtime system behavior.
%As opposed to static analysis of source code, traces examine the behavior of demonstrably feasible paths during runtime operating conditions.
This body includes the inference of invariants of different types \cite{Robillard2013AutomatedAP}, 
from those attempting to characterize a variable range of values \cite{Ernst2000QuicklyDR,daikon, Hangal:2002,JiangED13, NguyenDig2014}
%\se{add a few more state: Diduce} 
to those infering temporal invariants
\cite{Gabel2008JavertFA,Le2018DeepSM,Lorenzoli2008AutomaticGO,Nguyen2017SymlnferIP,Walkinshaw:2016,Ammons:2002, Yang:2006:PMT}.
The body of work also utilizes a variety of inference mechanisms ranging from  
%\se{what does daikon uses, how about dig}
frequentist inference \cite{daikon}
to the generation of polynomial relations \cite{NguyenDig2014}
to k-tail \cite{Lorenzoli2008AutomaticGO} 
%to symbolic traces \cite{NguyenDig2014}\se{Not sure this is an approach}
to deep learning \cite{Le2018DeepSM}. 
% to inference based upon symbolic states \cite{Nguyen2017SymlnferIP}.
%\se{make description above rich enough with a pointer to background for more details}  
As we explored this body of work for its application to 
autonomous mobile robots, however, we came to realize that 
these kinds of systems introduce a couple of unique 
attributes that existing invariant generation approaches were unable to fully capture.

%\ms{Reviewer 1: use of ``states'' is confusing. level of abstraction? Also, doesn't like that our approach ``assumes states are fully given'' 
%as opposed to inferring states like other approaches can do.}
%\se{I tried to clarify the notion of state in the examples. We do assume that a ``space'' of states are given, and we provide support to explore them systematically. 
%I did not think this was the right place to discuss that, but we need to say it later under contributions now. Take a closer look to the next few paragraphs that have changed}
%\se{revised paragraph - check}
The first unique attribute is the extent to which \textbf{different system states render distinct sets of invariants}, as the behavior of these systems is conditioned not so much by typical programmatic structures (i.e., functions' pre- and post-conditions),  but rather by particular system states.  
%These conditions are often modeled in finite state machines, which clearly subdivide the event space of the system into various system states.
For example, the sensors activated and the attitude of a drone are conditioned by different mission states such as takeoff, 
approaching a target, or tracking a target, while a self-driving vehicle's linear velocity bounds may change depending on whether the car is charging, parking, driving within a city, or driving on a highway. 
We argue and later show that ignoring this attribute greatly limits the potential of uncovering valuable invariants that only appear under certain states.

The second distinctive attribute is the \textbf{degree of uncertainty intrinsic to these systems}. 
They may render different results under the same environmental conditions due to sensor noise, 
imperfect estimators, inaccurate actuators, and humans in the loop. 
For example, the GPS sensor of the drone's localization component we later study 
%outdoor localization estimates based on unobstructed GPS 
has an accuracy of $\pm3.0$ meters,  the target recognition component success depends on the hovering stability which affects the camera's ability focus, 
%\footnote{\url{https://www.parrot.com/global/support/products/parrot-bebop-2/faq-bebop-2}},
%actuators like those associated with the PX4 flight stack are often capped 
%below their max to avoid actuator saturation\footnote{\url{https://dev.px4.io/%v1.9.0/en/concept/mixing.html}},
%and actuators in general are often oversold in terms of the electromotive force they output %\footnote{https://www.miniquadtestbench.com/motors}
and human operators with similar training can have a wide variety of reaction times.
%\se{provide another example of an actuatorm and then a human in the loop in terms of her responses to complete the picture}. This uncertainty implies that invariants characterizing these systems cannot be absolute but rather probabilistic.
We argue and later show that failing to handle this uncertainty attribute properly will make it difficult to  judge the value of an inferred invariant.

%\ms{Talk about repair techniques, such as invariant restoration. Complexity of CPS necessitates automation of these repairs.}
%\se{new paragraph - check}
\textbf{The state of the art in invariance inference, however, does not support the generation of invariants that are conditioned by arbitrarily complex system states, nor does it support probabilistic invariants to better characterize the uncertainty associated with the exposed behaviors.}
The closest related work considers having two outcomes happening jointly (that is, occurring at the same time, but not conditioned on each other) and ignores the prior system probabilities by making assumptions about the data distribution \cite{daikon}.

%conditional probabilities, and though some existing tools could be extended to do so
%\se{@meriel: which ones, how?}, they could not incorporate the continuous updating that recommends a Bayesian approach \se{@meriel: there is a gap here, we went from conditional probabilities to bayes before introducing bayes}. \ms{Not certain I can say the current state of the art doesn't handle conditional probabilties. \cite{Le2018DeepSM} approximates conditional probabilities using a statistical language model}
%\ms{What do we suggest and how is it useful? How is it novel (novel in application)? What makes events that exhibit these statistical characteristics useful/interesting?}

 In this work we address this challenge by building on the statistical structure of conditional probabilities, $Probability(Outcome|Given)$, to uncover likely invariants that only manifest under particular given program states.  As we shall see, our approach generates invariants    such as:

 \vspace{-0.12in}
{\footnotesize
\begin{equation*} 
\begin{split}
P(applied Throtle& = 0 \mid   \\ 
& vehicleMode=autonomous \quad \wedge  \\ & pedestrianLocation=roadway)>0.99 \\ 
\end{split}
\end{equation*} 
} 
\vspace{-0.12in}

which indicates that, given that a vehicle is operating in autonomous mode and a pedestrian is closeby, the probability for the throttle to be disengaged is over 99\% (in contrast, without such conditioning, we could only learn that the Throttle range is between 0 and 100).
%a reconnaisance drone such as:
%{\footnotesize
%\begin{equation*} 
%\begin{split}
%P(targetSensor & = partialTargetSensed \mid  \\ 
%& missionState=targetDetected \quad \wedge \\ & velocity=low \quad \wedge 
%\\ & userCommand=none
%)>0.3 \\ 
%\end{split}
%\end{equation*} 
%}
%which indicates that, given a drone perfoming a missiong that has moved into the target detection stage and that is moving with a low velocity, the probability that a target is partially sensed is over 30\%. 

\textbf{Problem requirements.} 
We aim to fulfill four requirements to make the approach practical. 
1) Provided with   a high-level specification of the potential variables to explore as part of the $Outcome$ and $Given$, the approach must systematically investigate the space of relevant predicates to those variables as part of the $Outcomes$ that   only hold under predicates on those variables as part of the $Given$. Note that such high-level specifications can be provided by an engineer or by existing invariant generation tools like the ones cited earlier.
2) The approach must enable the developer to uncover valuable conditional invariants among the large space of predicates explored.   
%without overfitting. Overfitting presents a challenge in that the indiscriminate addition of predicates into the $Given$ may render invariants that hold with high probability but are applicable to a very small number of potential executions, and hence deemed less valuable.
3) The approach must be able to incrementally incorporate prior knowledge as it becomes available, either from a new collected trace or from a developer's knowledge, to improve the probability estimates, without incurring in the cost of recomputing all invariants when new  data is added. This is particularly important as multiple sources of evidence become available as the system evolves.
4) The approach must avoid relying on arbitrary thresholds 
to determine what is and is not significant as the choice of such thresholds is highly dependent on the context.

\textbf{Conceptual Solution.} To address the first requirement, we define a family of initial relevant predicates patterns for the robotics domain and 
a Bayesian invariant inference engine that implements conditional inference, and prototype
a domain-specific specification language  and a tool pipeline to compute them.
To address the second requirement, we incorporate   
ranking mechanisms that assist in judging an invariant value based on how much an invariant probability changed from prior estimates to posterior findings.
% and that uses an information content metric to select the invariant per outcome that offers best fit with the least parameters to avoid overfitting. 
To address the third and fourth requirement, and also to further support the first requirement, we shift the inference model from using the classical (frequentist) statistics employed 
by existing approaches (e.g., \cite{daikon,Hangal:2002, JiangED13, Yang:2006:PMT}), to a Bayesian inference model that allows us to easily incorporate prior information from previous traces or developer's knowledge, and  does not require the definition of arbitrary thesholds or the reliance on data distribution assumptions.

 The contributions of this work are:
\begin{itemize} [nosep]
	%\item The addition of Bayesian conditional invariants as an invariant type to augment the existing field of invariant generation techniques. %\se{maybe something about problem definition, or recognition of missing elements in invariants}
	\item Approach to infer a family of conditional invariants   from traces through Bayesian inference, and mechanism to rank the solutions based on probabilities and surprise
	%\se{add one more sentence: why is great? unique?} 
	%to capture and continually increase the accuracy of probabilistic distributions of 
	%conditional behaviors.
	\item Implementation  that provides the mechanisms to specify the space of variable predicates to explore, and launch the inference engine to systematically explore that space.\footnote{\url{https://anonymous.4open.science/r/838f6ec4-c4c4-4ce7-a7c4-8910c3a73e66/}}
	\item Assessment of the proposed approach through its application to two systems, a reconnaissance drone and a semi-autonomous simulated car. The findings indicate that the approach can uncover valuable invariants that cannot be generated by existing approaches.
	%The assessment showcases the valuable and distinct body of bayesian conditional invariants that can be generated through the proposed approach, and the tradeoffs between costs and benefits associated with the scope of invariants explored.  

\end{itemize}
 
%\se{@Meriel: please update the github repo with the latest tool and data}

\section{MOTIVATING EXAMPLE}
 
\begin{figure}
	\centering
	\includegraphics[width=2in]{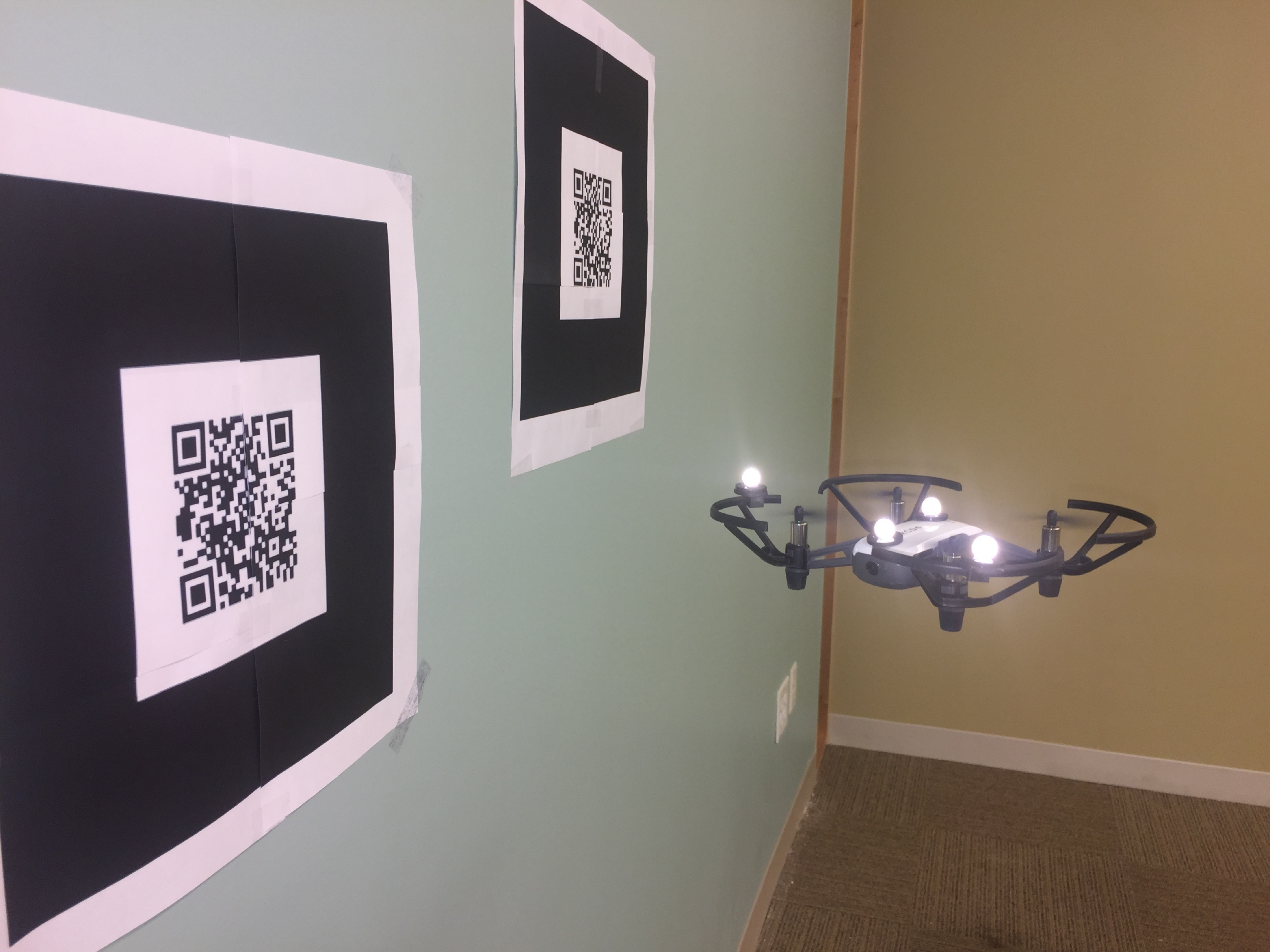}
 	\caption{Drone ISR scenario.}
	\vspace{-0.1in}
	%\caption{DJI Tello inspects potential target during drone ISR scenario.}
	\label{fig:drone-scenario}
\end{figure}

Consider a drone performing a surveillance mission over a field whose objective is to locate and confirm a target. 
Figure \ref{fig:drone-scenario} shows a downsized version of this real-world scenario where a micro-drone is sweeping for QR code targets.% posted on a wall inside of a user-defined area. 
  
%\se{@Meriel: please consider updating tables of first study to better match change states names used in here to make them more readable. It helps to separate more clearly machine vs mission, and use mission 'phase' instead of 'state' to avoid overloading 'state'}

This system contains a \textbf{rich set of states that condition system behavior.}
The system state includes a large set of variables-value pairs, from the drone's pose to its mission phase. 
Existing techniques \cite{daikon, Gabel2008JavertFA} can readily capture state invariants 
such as $0.0<droneAltitude<400$ as preconditions when the drone enters its navigation 
subroutine, or even as temporal invariants such as $\boxempty((dronePhase=Takeoff) \implies  \diamond (dronePhase=Land))$. 

Existing techniques, however, fail to explore the rich and valuable space of invariants that 
are conditioned by system states. 
For example, although drone altitude can change over the range of a mission, 
when landed, the relative drone altitude should be zero. More specifically, we want to capture the conditioned drone altitude by the phase as in $(droneAltitude=0 \mid dronePhase=Land)$, effectively refining the earlier invariant on $droneAltitude$  as it is conditioned by $dronePhase$.
In other cases, not just refined but totally new invariants emerge as the likelihood of the Outcome 
increases enough under a particular condition to be reportable. 
For example, since a drone's altitude can change significantly while flying and following a trajectory, no invariants on altitude changes would be usually reported as there is no sufficient evidence to support an inference. However, the altitude of a drone should remain fairly consistent  while hovering, so  the $dronePhase$ conditions the altitude change as in 
$(|Alt_{start:hov} - Alt_{end:hov}| <  \delta \mid dronePhase=Hovering)$.

Most of the mentioned   \textbf{states are coupled with sources of stochasticism}, such as sensors affected by environmental conditions (lighting affecting target recognition capabilities), fusing estimators influenced by sensor noise (accuracy of the reported gps and pressure sensor altitude), or users variability (mission commands that are inconsistent  even under the same context). As such, the conditioned and the conditioning states  are inherinetly uncertain.
So, while it may be correct to report that  $0 \leq droneSpeed \leq 5$,  we would like to produce a conditional characterization that relates velocity and target recognition
such as $(targetSensor=targetDetected \mid 0 \leq droneSpeed \leq upperSpeed\wedge missionPhase=targetDetected)$. However, such invariant would not be that useful because for it to hold with total certainty $upperSpeed \approx 0$, which reduces the conditioning space so much that the value of the invariant decreases as it will only hold very rarely. Instead, what we envision is to be able to include larger portions of the state for invariants to be more useful but accommodate the uncertainty as part of a probability to form our proposed structure of $P(Outcome \mid Given)$. That would help us generate $P(targetSensor=targetDetected \mid 0 \leq droneSpeed \leq upperSpeed=3 \wedge missionPhase=targetDetected)>95$ which indicates that there is a high likelihood, over 95\%, that when the drone velocity is under 3 m/s and the mission phase is target detected, the target will be sensed properly.  One more distinguishing feature of the proposed approach is that to compute (and subsequently update) the probabilistic distributions associated with these outcomes, we use Bayesian inference to determine the conditional probability. As we shall see, this approach let us incorporate new traces into the inference process more efficiently and it avoids the use of artificial thresholds for reporting.

\section{Related Work}

As mentioned in Section \ref{sec:introduction}, our community contribution to invariant generation has been extensive. We now decribe in more detail the work that most influenced  our effort.

Ernst et. al. \cite{daikon} established Daikon, one of the benchmark inference engines for detecting program invariants. 
Daikon's engine creates a field of potential invariants based on a set of predefined invariant templates and a trace. We follow a similar approach in that our predicates are basic patterns, but we incorporate them in the richer conditional probability structure. Daikon then evaluates the potential invariants 
according to whether there are sufficient samples to support them and no samples that violate them. 
This frequentist approach, prevalent among invariant inference engines, uses a confidence interval to ascertain that a predicate holds against some probability of random negation, determined by the number of samples supporting that predicate. 
%\ms{Reviewer 1: be more specific about what that significant cost is. Algorithmic complexity? Labelling? Number of traces?}
One drawback of their frequentist approach is that, while new traces can be added to an existing sufficiently large set of samples, there is a significant cost associated with trace accumulation and algorithmic complexity needed to reach that sufficiently large set without relaxing the confidence interval. 
%\se{careful, not true, it can add additional traces, it just does not build on prior knowledge though - revise} 
Furthermore, although Daikon computes joint probabilities, it does not compute conditional invariants, nor multi-predicate joint probabilities.
%To introduce a conditional invariant into the Daikon invariant templates would be difficult to incorporate into a strongly typed template schema of Daikon, would require manual definition by the developer, and would be limited first to equalities and then ranges with additional adjustments.
%\ms{Not sure why Daikon wouldn't support conditions between ranges, as it does support unconditioned ranges. Definitely couldn't support windows of values.}
%\se{Also need to say that Daikon does introduce conditional, must they must be manually defined by the developer, but can only support equality, not ranges, right?}
%The daikon tool follows a generate-and-check approach in that all possible invariants are generated. Weaker, low-confidence invariants are suppressed and invalidated invariants are discarded. Invariants that are always simultaneously true are presented as implications (if-then invariants). 

%\ms{Metric temporal logic intro'd here. }
Perracotta \cite{Yang:2006:PMT}  extracts temporal API specifications from traces through a mix of analyses, patterns, and heuristics. Perracota was among the first to incorporate mechanisms to 
ignore potential blips of  noisy trace content  as negligible in the context of the overall trend of behavior. 
Our search for probabilistic invariant generation is inspired by these kinds of challenges.
In a similar line of work, Gabel et al. \cite{Gabel2008JavertFA} developed a  mining framework of 
temporal logic properties. 
Their approach is similar to many approaches in terms of combining patterns and 
incrementally encoding them as FSMs. However, their strategy to start with simple patterns
that can be composed to generate much more complex ones is one that we have adopted in our approach.
%template handling to ours, in that property templates are generalized through a set of simple predicates that can be composed into more complex property specifications. These patterns can be iteratively subsumed into higher-level macropatterns of temporal behavior, much like the con/disjunctions of and proposed extensions to predicates in our engine. Ammons et al. \cite{Ammons:2002} approaches temporal specification mining by applying machine learning to an assumed-correct program execution. Specification precision proved challenging to their automaton learner, which we have characterized as an issue of domain knowledge for the developer. 

Grunkse \cite{grunske2008} focuses on probabilistic invariants as a qualitative expression of requirements. 
He introduces a rich set of specification patterns coupled with a structured English grammar to express bounded 
probabilistic behavior of a system, instead of in terms of absolute correctness, to incorporate expert knowledge 
and for it to  be used for formal verification. Although this work did not pursue automated inference of invariants,
its treatment of probabilistic patterns offers a roadmap for us to expand our work.
%Similarly, Le et. al \cite{Le2018DeepSM} use deep learning techniques to improve probabilistic temporal specification model accuracy.

Jiang et al. \cite{JiangED13, jiang:2017} extend the Daikon invariant library to patterns seen in robotic systems in order to derive 
monitors that can check system properties at runtime. While Jiang et al. introduce invariant templates tailored to robotic systems (e.g., deployed processed and their relationship,
bounded time differentials, polygonal relationships between spatial variables), their approach still relies on a traditional frequentist approach and does not considers conditionals. Aliabadi et al. \cite{artinali} present a similar approach for cyberphysical system security with a focus on reduction of false positives and false negatives.
%\se{@Meriel: say a bit more about their approach, a couple of more lines}

%, and so all 83 of their test runs must be processed simultaneously.
%Lorenzoli et al. \cite{Lorenzoli2008AutomaticGO}, another traces-only inference technique, employs the k-tail inference 
%algorithm to generate system models.
%\ms{Commented out \cite{Lorenzoli2008AutomaticGO} mention; mentioned in intro}
%Nguyen et al. \cite{NguyenDig2014} created the Daikon-adjacent DIG tool to derive novel nonlinear numeric state invariants from dynamic program traces.

%\se{I feel that we should drop this as they are not closely aligned to others. Let me know if you feel that is not the case}
%Krka et al. \cite{Krka2014AutomaticMO} evaluate the quality of several FSM generation tools, some from our related works \cite{Ernst2000QuicklyDR, Gabel2008JavertFA,Lorenzoli2008AutomaticGO,Yang:2006:PMT}, according to the sources of information used by the specification mining techniques.
%,  and summarize by saying using filtered invariants provides an initial higher-quality FSM model of a program, and invariants and traces are needed to infer a robust model.
%Robillard et al. \cite{Robillard2013AutomatedAP} characterize the output of API property inference techniques as mentioned in the introduction.

We note that none of the approaches, although closely aligned with ours, produce the conditional probabilistic invariants with Bayesian inference that we are pursuing.

%Worth mentioning:

%Ammons et al. \cite{Ammons:2002} performs temporal specification mining by applying machine learning to an assumed-correct program execution. Specification precision proved challenging to their automaton learner, which we have characterized as an issue of domain knowledge for the developer to decide upon configuration of the engine. 
%Similarly, Le et. al \cite{Le2018DeepSM} use deep learning techniques to improve specification model accuracy.

% symInfer: derive symbolic states from symbolic program executions and uses counterexamples to winnow down the candidate invariants to a set that describe program behavior. While it proves itself a powerful tool, it is limited to numerical variables.

%\ms{introduce concept of metric temporal logic (see approach overview)}

\section{APPROACH}
%\ms{Change all ``A events'' to ``outcome''.Don't oversell complexity of Bayesian inference technique. Mention compound given predicates where it will complicate the approach, which is currently demonstrated on a single given predicate.}

%\se{Something was off with the terminology -- we use events vs states. Revise for consistency again}
The goal of the proposed approach is to generate invariants that 
capture the probabilistic influence between system states as in $P(Outcome|Given)$.  The next sections describe how, by building on conditional probabilities 
and Bayesian statistical inference, we can process system traces to 
produce invariants that meet that goal.

\subsection{Building Blocks}

%\se{@Meriel: regenerate Figure 1: Remove "model Selection" from the last box, andd add the word "Invariant" at the top of "Posterior Probabilities"}

We capture the probabilistic influence between two dependent states as conditional probabilities of the form $P(Outcome \mid Given)$,  where $Outcome$ and $Given$ are \textit{boolean predicates evaluated over single or multiple states of a system}.  
This simple conditional pattern belies the richness of invariants it can encode, as these predicates can take many forms and 
can be composed to form arbitrarily complex descriptions of   states.

In its simplest form, if $Given$ is TRUE, then $P(Outcome \mid Given) = P(Outcome)$. 
This implies that any existing invariant patterns explored in the related work or generated by existing invariants toolsets (e.g., \cite{daikon, Nguyen2017SymlnferIP, JiangED13}) can be subsumed by the proposed this encoding. 
That includes from the simplest form of state invariants such as $battery > 0$ 
to complex metric temporal logic formulas such as $\boxempty(Takeoff) \implies  \diamond_{[0,3]} (altitude>0))$. 
We also later discuss in Section \ref{sec:implementation} the patterns that we currently support in our own infrastructure, whose selection was driven in part by the needs we observed in the robotic systems  we have developed and studied. 
%For the subsequent explanations it suffices to state that both $Outcome$ and $Given$ consist of predicates on single or multiple sequence states.
%where $\Box$ and $\Diamond$ denote the \emph{always} and \emph{eventually} temporal operators, respectively \cite{alur1991logics}.
%\lu{@Meriel, please cite a reference of MTL paper here if it hasn't been mentioned before.} 
%\lu{has metric temporal logic been introduced somewhere in the paper? Where is the definition of $\Box$ and $\Diamond$?} \ms{metric temporal logic will be introduced as a concept in motivation. Are these short definitions sufficient?} \lu{I edited the last sentence of explaining $\Box$ and $\Diamond$}

In its richest form and the focus of this effort, $P(Outcome \mid Given)$ lets us explore how the system state encoded in $Given$ conditions the occurance of other  states  encoded in $Outcome$. 
%For example, continuing with the drone scenario described in the motivation, given
%$A:TargetDetected=TRUE$ and $B:VelocityHigh=TRUE$, we could consider the conditional probability  \textit{P(TargetDetected $\mid$ VelocityHigh)}
%\footnote{For readability, we simplify the predicates that check a boolean variable by just specifying the variable name}, 
%which describes the probability of a drone sensor detecting features of a target, given that the drone is moving at high speeds.

\subsection{Bayesian Inference}

\begin{figure}
	\includegraphics[width=0.9\columnwidth, scale=0.25]{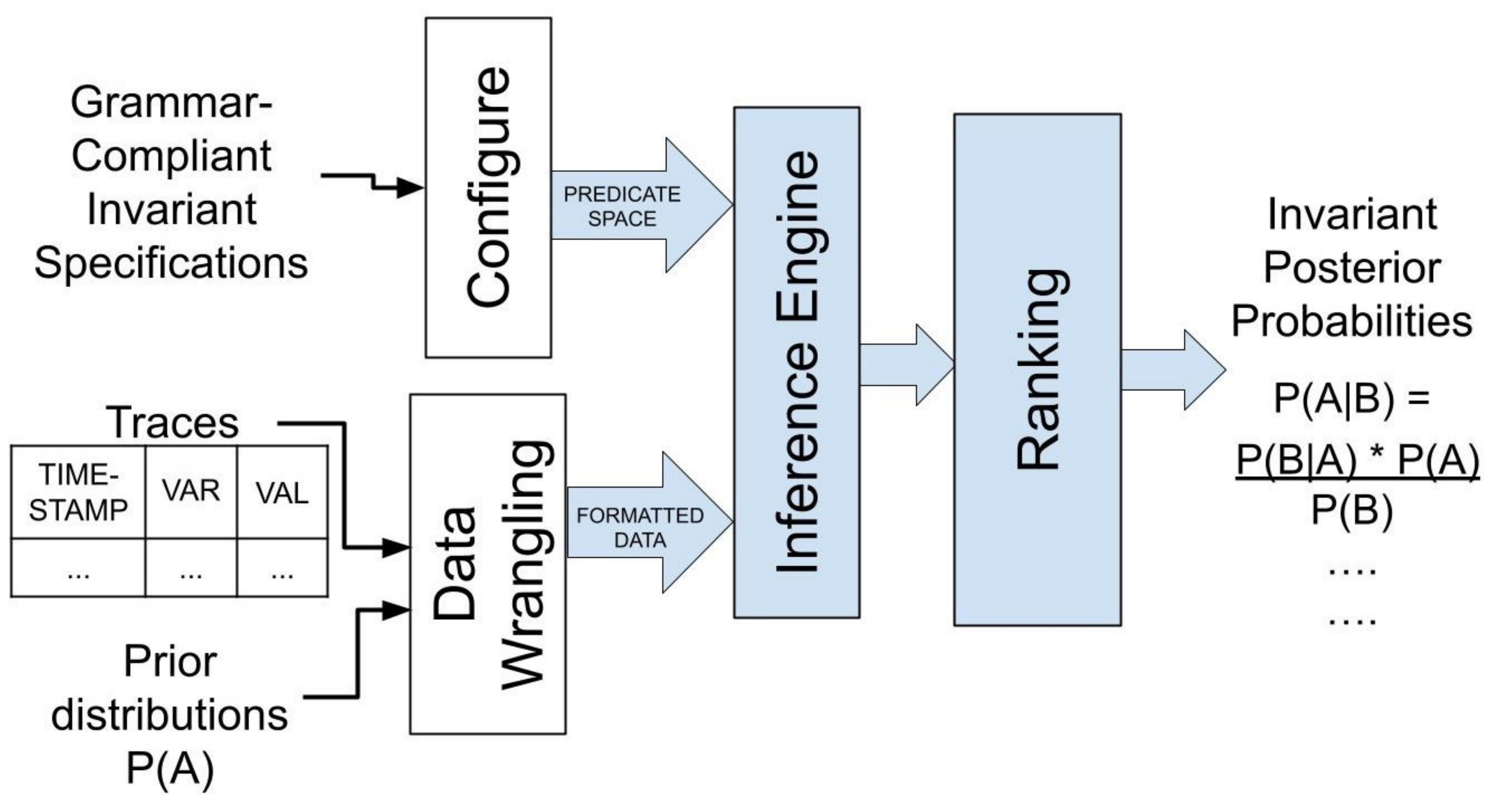}
	\vspace{-6mm}
	\caption{Approach Overview.}
	\label{fig:engine-diagram}
 \end{figure}

The inference engine that computes $P(Outcome \mid Given)$ takes as input a space of boolean predicates to explore as part of $Outcome$ and $Given$, an execution trace of time-stamped variable-value pairs, and  set of prior distributions corresponding to the prior probabilities which reflect whatever knowledge the developer has on the $Outcome$. This engine sits in the middle of a larger framework depicted in Figure \ref{fig:engine-diagram} that we have prototyped and that is described in more detail in Section \ref{sec:implementation}. Briefly, the configure module checks the specification of the space of predicates to explore for completeness, provides a summary of the $Outcome$ and $Given$ space to be explored, and it generates a configuration to guide the inference engine in the analysis of the traces as per those predicates. The data wrangling module preprocesses the trace to put it in the right engine format. The configuration and wrangling modules are described later in more detail.

The inference engine relies on Bayesian inference to compute conditional probabilities based on the predicate space and information in the trace and the prior.   The engine produces such probability as per the Bayesian formula:

\[P(Outcome \mid Given ) = \frac{P(Given \mid Outcome) P(Outcome)}{P(Given)}\] 

where $P(Given \mid Outcome)$ and $P(Given)$ are computed by the engine from the trace, while $P(Outcome)$ comes from the priors. 
In the next section we will discuss how the generated conditional probabilities can be ranked by the posterior probabilities or by  the change  from the prior.

%\se{@Meriel: please check that we do not have any A and B but instead O and G in this section}
 
Algorithm \ref{alg:process-trace} shows the key steps in the inference process for two predicates, $O$ in the $Outcome$ and $G$ in the $Given$, to form $P(O \mid G)$.
%Note that we discuss its generalization and optimization for handling multiple predicates as part of $P(A|B, C, ...)$ in Section \ref{sec:implementation} and more broadly in Section \ref{sec:conclusion}. 
Given a trace, those two predicates $O$ and $G$, and a prior distribution for $O$, the algorithm starts 
by processing each record in the trace, and evaluating the predicates on the appropriate trace variables.
Such evaluation is performed according to the type of predicate to render a \textit{TRUE} if the predicate holds.
If predicates $O$ or $G$ hold, then their corresponding frequency count is updated, and if 
both of them hold then their joint probability is also updated. 
Once the trace is processed, these frequency counts are used to compute the probabilities required by 
Bayesian inference: the total probability P($G$) and the joined probablity P($G \mid O$).

The prior $P(O)$ that is provided as input to the engine can be obtained from sources such as previous deployments of the system in distint contexts, or tests results collected on variants of the system.
%\ms{Added sentences and equations about priors derivation:}
In our studies we derive the initial priors in one of two ways: either by assuming a uniform distribution, where all values of a variable are ascribed the same probability: \emph{X $\sim$ Uni[a, b]: p(x)= $\frac{1}{b-a+1}$}; or by using a disjoint subset of data as a sampling distribution arrived at by counting the observations of a certain variable value over observations of all values of that variable: \emph{X $\sim$ F$_x(x_i)$ = $\frac{freq(x_i)}{\sum freq(x)}$}.
%\se{@Meriel:  Can we bring a formula here and explain it? }
Equipped with those three probabilities, the algorithm can now finally compute  P($O \mid G$). 
Note that as defined,  Algorithm \ref{alg:process-trace} operates on a single trace. In practice,  building on Bayes updating, this algorithm is repeatedly invoked as new traces are collected and beliefs are refined, so that the information resulting from an analysis can be refined as more information is gathered.
%in a couple of different ways to illustrate the approach.
%In the first study we split the trace dataset to mimic traces collected in-house versus traces collected in the field, while in the second study we take one trace at a time mimicking new traces regularly coming from the field that provide updated information.
%That is, given a trace $t_i$ that resulted in $post_i = P(Outcome \mid Given)$, the analysis of $t_{i+1}$ will take into account $prior_{i+1} = post_i$, leading to an incremental adjustment of the invariants as new traces are analyzed. This is something we later explore in our second study.
%\se{What is in this paragraph needs to become part of the previous subsection -- see comment before. In here what we need is an example, so you could just explain what would happen if a new Prio arises like 0.2 for the same predicates, then what?}
To accomodate new data that may affect the initial computed priors, we have incorporated a priors update step in between traces. 
%Thus the prior for $missionPhase=targetDetected$ is updated as new information surfaces.
The number of observations for each $O$ predicate and number of timesteps in the previous trace, as well as the total number of timesteps from which the previous prior was generated are propagated forward for a frequentist update of the prior probabilities in which these probability distributions of the ongoing prior and the events in the previous trace are combined by weighting the supporting observations as such:
$$
Prior_{new} = \frac{Prior_{old} * timesteps_{old} + observations_{new}}{timesteps_{old} + timesteps_{new}}
$$
This allows us to leverage new trace data to build a more representative prior distribution, providing increasingly accurate invariant probabilities as new trace data is consumed.

Two additional aspects of the algorithm acquire additional complexity. First, the $eval$ function must deal with predicates that require processing multiple trace records concurrently, peeking backward and forward in the trace to evaluate  predicates over multiple states such as those describing a trend of values or a temporal relationship. For example, a predicate to determine whether the $acceleration$ is increasing within a time window, or a predicate that checks whether the $altitude$ eventually converges requires an analysis that spans multiple states. 
Second, compound predicates impose  additional frequency tracking, and extended functions to compute the probabilities as they require to iterate over a larger number of combinations of predicates. We  illustrate these challenges through an example in the next section.

\begin{algorithm}
	\SetAlgoLined
	\caption{Single Inference}\label{alg:process-trace}
	\KwIn{trace, $O$, $G$, prior P($O$)}
	\KwOut{P($O \mid G)$}
	% New definitions
	\SetKwProg{switch}{switch}{:}{endswitch}
	\SetKwProg{case}{case}{:}{}
	\ForEach{record in trace}{ \label{call:being-outer-loop}
		%		$evalA$ $\leftarrow$ eval($A$)\;
		%	    $evalB$ $\leftarrow$ eval($B$)\;
		\If{eval($O$)}{
			$freqO$ $\leftarrow$ $freqO$ + 1\;
			\If{eval($G$)}{
				$freqOandG$ $\leftarrow$ $freqOandG$ + 1\;
			}
		}
	}   
	P($G \mid O$) $\leftarrow$  $freqOandG$ / $freqO$ \;
	P($G) \leftarrow$ P($G \mid O) * $P($O) + $P($G \mid \neg O) * $P$(\neg O)$ \;
	P($O \mid G$) $\leftarrow$ P$(G\mid O) * $P$(O) / $P$(G)$\;
\end{algorithm}

\subsection{Applying Bayesian Inference}

%In practice, a developer may want to explore a large number of predicates as part of the $Outcome$ and the $Given$, which is why we provide some support to specify the exploration space through a specification grammar and some predefined patterns (see Section \ref{sec:implementation}). 

For simplicity, let us assume a developer is interested in  just exploring a space that includes predicates $O: missionPhase=targetDetected$ and  $G:  velocityChange[3] < 0$
(where the 3 corresponds to the number of timesteps back to consider the change).
%and  its ranges are $y\mhyphen velocity < 0$, $y\mhyphen velocity \geq 0$. 
%This is just for illustration as 
The developer also provides, based on knowledge earned through previous deployments, a $missionPhase=targetDetected$ prior of $0.3$ representing the likelihood of the drone state to be in that discrete state at any time during the execution.
Finally the developer provides the brief sample trace in Table~\ref{table:sample-trace}.

\begin{table}
	\caption{Simplified trace and predicates evaluated on the drone scenario.}
 	\label{table:sample-trace}
	{ 
	\centering
	\begin{tabular}{l l r | c c }
		\hline
		\multicolumn{3}{c|}{\textbf{Trace}}	& \textbf{Outcome} & \textbf{Given} \\
		\textbf{T} & missionPhase & y-vel  & missionPhase &  velChange[3]\\ 
		  &   &    & =TargetDet. & $<0$\\ \hline
		  n &  ... & ...& &  \\ \hline
 		n+1    &  Sweeping     & 0.1   & &   \\ \hline
		n+2    &	Sweeping	 & 0.2  & &   \\ \hline
		n+3    & Sweeping  & 0.05    & &   True (0.05$<$0.1)\\ \hline
		n+4    & TargetDet.  & 0.0  & True &  True (0.0$<$0.2) \\ \hline
		n+5    & TargetDet.  & 0.1  & True &   \\ \hline
		n+6 &  ... & ...&  & \\ \hline
	\end{tabular}
	}
\end{table}

Algorithm \ref{alg:process-trace}  processes the trace to compute the frequency  of $missionPhase=targetDetected$ and of the likelihood $missionPhase=targetDetected$ given $velChange<0$. 
Since the velocity window is 3 and ignoring windows extending outside the shown trace, the algorithm counts two instances where velocity decreases: from times 1-3 and from times 2-4. 
Times 4 and 5 have an instance of $missionPhase=targetDetected$.
So, the frequency counts after processing the trace are: 
$missionPhase=targetDetected$: count = 2, 
$ velChange<0$: count = 2, and 
$ velChange<0 \mid missionPhase=targetDetected$: count = 1.

After the trace is processed, in line 10,   $P(G)$ is calculated as per the law of total probability: 
$P(G) = \sum_{i=1}^{n} P(G\mid O_i)P(O_i)$, 
where $O_i$ denotes a value of predicate $O$ and $P(O_i)$ is the prior probability of $O_i$. In our example, since we are working on just a single predicate $O$, $O_1$ corresponds to $missionPhase=targetDetected$, and we use the complement 
$\neg missionPhase=targetDetected$ as $O_2$ to compute the total probability.
(If another predicate like $missionPhase=Landing$ was defined by the developer, 
then that would constitute the new  $O_2$, and the complement of those predicates' conjunction would be $O_3$.) The prior of $missionPhase=targetDetected$ is 0.3, so the prior of $\neg missionPhase= targetDetected$  is 0.7. Then, since $missionPhase=targetDetected$ occurs 1 out of the 2 times that $velChange<0$ is observed and 1 out of the 3 times $velChange \geq 0$ is observed,
then we have $\frac{1}{2} * 0.3 + \frac{1}{3} * 0.7 = 0.383$ as the total probability of $missionPhase=targetDetected$.

The prior, likelihood, and total probabilities are 
used to calculate the Bayesian probabilities as: $P(missionPhase$=$targetDetected \mid  velChange<0)=$
$\frac{\frac{1}{2} * 0.3}{0.38}=0.39$. 
%In a full run, $P(y$-$velocityChange<0 \mid$ $MachineState$$\ne$ \\$PossibleTargetDetected)$, $P( y\mhyphen velocityChange\geq 0 \mid$$MachineState$=\\$PossibleTargetDetected )$, and $P( y\mhyphen velocityChange\geq 0 \mid$$MachineState$$\ne$\\$PossibleTargetDetected )$ would also be computed.

%\se{@meriel: could you add a few lines of what would take for the algorithm to deal with an invariant like P(A|B,C) with an example? Just how many invariants would have to be explroed for 3 predicates, and what would it take to compute P(B,C) in terms of additional terms (like what would line 10 look like in number of terms?)}

This illustrative example focuses on predicates with expressions on just single variables. 
%The reason is twofold. 
%First, there are many valuable invariant patterns we have encountered in autonomous systems that take this simple form and that can serve to validate the viability of the proposed approach.
However, it can be easily generalized to handle multiple given predicates, where the algorithm checks for the coincidence of $Outcome$ and multiple $Given$ predicates instead of a single one. So, given predicates $O, G1,$ and $G2$, to calculate $P(O \mid G1, G2)$, $freqAandG1andG2$ is  computed to produce $P(G1, G2 \mid O)$ on line 9 in Algorithm \ref{alg:process-trace}, and line 10 would become  
$P(G1,G2) \leftarrow P(G1, G2 \mid O) * P(O) + P(G1, G2 \mid \neg O) * P(\neg O)$.

The  inference engine implementation, available here\footnote{https://github.com/MissMeriel/inference\_engine},
generalizes Algorithm \ref{alg:process-trace} to accommodate multiple predicates, performing 
additional bookkeeping  to optimize the evaluation of those predicates, requiring only one traversal of the trace.  
The implementation uses the Apache Commons Lang \cite{commons:lang} 
and Apache Commons Math \cite{commons:math} packages to process original and intermediate trace files.

\subsection{Invariant Rankings}\label{rank-filter}

As the potential space of predicates to explore grows, so does the size of the potential space of invariants the approach must evaluate, and the ones a developer must analyze. 
As such, we found it  essential to integrate mechanisms to highlight invariants that may of be of value to the developer.

%\se{Reviewer's concern: The rank of a conditional invariant increases as P(Outcome) gets smaller and P(Outcome|Condition) gets larger. In other words, a conditional invariant has a high rank if the Outcome is rarely observed overall, but whenever it is observed, it is mostly observed together with the Givens.}

The product of the inference engine, $P(Outcome \mid Given)$, offers a natural initial ranking criterion.   The probabilities generated by the inference engine constitute a simple and effective way to rank invariants as per their likelihood.

However, for a context with an evolving system and changing deployment environments where  new evidence and traces will re-trigger the invariant inference process frequently, the most valuable nugget of information is on the invariants that offer the most change. More specifically, we are interested in how much the latest invariant findings  surprise the developer. For this, we build on the surprise ratio \cite{IttiBaldi06nips}, expressed as the posterior likelihood over prior probability of the outcome:
 \[Surprise(Outcome) = P(Outcome \mid Given) / P_{prior}(Outcome)\]
Surprise ratios can be interpreted as the factor by which an outcome becomes more likely in a conditioned state. An invariant $P(velocityHigh \mid batteryLow)$ with a  surprise ratio of 2 indicates that the outcome  $velocityHigh$ is 2 times more likely when conditioned upon the given $batteryLow$. If the prior and posterior are identical, then there is no surprise and the ratio is 1. 
%\se{I went back to the original papers and found some differences with what we say here: http://ilab.usc.edu/publications/doc/Baldi_Itti10nn.pdf
%https://www.cc.gatech.edu/~dellaert/dhtml/pubs/Ranganathan09icra.pdf
%A different version shown is this:
%\[ =P_{prior}(Outcome) *  log ( P(Outcome \mid Given) ) / P_{prior}(Outcome) \]
%So the log makes it closer to information content which makes sense, 
%and the new multiplier term would make having a smaller  prior-Outcome as a multiplier   rank those low prob invariants lower.. Thoughts?
%}

%\se{@Meriel: let's talk about this one. Need to mention the surprise ratio variance here?}
%\ms{added sentence about variance.}
Just like the computation of the Bayes estimates, the computation of surprise using the above equation is recursive as the posterior in one epoch consumes the prior for the subsequent step.  %This definition of surprise implies that an invariant that is surprising at first, when is observed repeatedly, it loses its surprising nature.
If both the prior and posterior of an invariant stabilize as more traces are added, then the surprise ratio stabilizes as well and variance of the surprise ratio decreases over time.

%Because the surprise ratio naturally becomes higher for rarer events  in the trace but are highly correlated with other events, some consideration is given to the number of observations of this event. A lower number of observations in the trace may mean this event is anomalous, an artifact of data cleaning, or due to data from a state persisting past the time that that state has been left. This is discussed further in Section \ref{sec:auto-driving-study}:Autonomous Driving. 

%
%\se{need more, will we just get the hyper-specialized patterns?}
In addition, to control for overfitting, we rely on the Bayesian information criterion (BIC) \cite{wit2012}. This metric maximizes log likelihood while penalizing overfitting of the model through the formula $BIC = ln(n)k - 2 ln(\hat{L})$, where $n$ is the number of observations, $k$ is the number of model parameters, and $\hat{L}$ is the maximum log likelihood.
% \se{@Meriel: add formula}.
Given two invariants with the same outcome, BIC will determine whether refining an invariant with more predicates is worth it. For example, if \textit{P(WarningState=BatteryLow $\mid$ x-velocityHigh)} contains \textit{P(WarningState=BatteryLow $\mid$ x-velocityHigh y-velocityHigh)} and \textit{P(WarningState=BatteryLow $\mid$ x-velocityHigh MissionState=Complete)} and the nested models with more predicates show little fluctuation in the resulting likelihood of \textit{WarningState=BatteryLow}, then the BIC will assign a more favorable score to the first model, the one that is the simplest and contains  the least predicates.
%\se{Need to finish sentence explainig how BIC would help}

\subsection{Configuring the Specification Space}
\label{sec:implementation}

%The portion of the implementation that requires more detail, however, are the patterns that are currently supported by the framework and the initial grammar to support a developer in specifying those patterns. The implementation provides a built-in family of specification patterns. The  pattern selection was guided in part by the needs we observed in the robotic systems  
The space of predicates to be explored as part of the $Outcome$ and $Given$ by the inference engine can be   specified by a developer or another invariant generation engine (we used Daikon for example in one of our studies) through our  specification grammar.  The grammar, abbreviated for space constraints in Listing \ref{lst:grammar}, was implemented using the LALR-1 CUP parser generator \cite{cup}. The grammar enables of variables to be included in predicates, of variables to be considered in particular predicates, and of 
constraints that allow the engine to prune the problem space by specifying what predicates should be considered as part of the $Outcome$ and $Given$. 

%For example, if the specification file defines the $Outcome$ variables \textit{x-velocity} and \textit{y-velocity}, and the $Given$ variables \textit{MissionState} and \textit{WarningState}, then adding the constraint \textit{P(x-velocity $\mid$ MissionState)} will condition \textit{x-velocity} only by \textit{MissionState}, as opposed to applying the predicates of all $Given$ variables to the predicates of the \textit{x-velocity} $Outcome$ variable. 

The semantics corresponding to the grammar already provide support for specific predicates types for which we have developed support in the form of $eval$ functions for Algorithm \ref{alg:process-trace}. The selection of these predicates for implementation was driven by the needs of the target domain and the follow up study and include three basic types of built-in predicate patterns: $Equality$, $Range$, and $Trend$.  

\begin{footnotesize}
	\begin{lstlisting}[linewidth=\columnwidth,breaklines=true, caption={Specification Grammar}\label{lst:grammar} ]
	<start> ::= OUTCOMES <pred-def>* GIVENS <pred-def>* CONSTRAINTS <constraint_def>*
	 <pred-def> ::=  <var name> ',' <type> ','  <threshold> ',' 
	 <partitions>  ',' <window>
	 <constraint_def> ::= 'P(' <var_name> '|' <var_name>* ')'
	 <type> ::= `INT-Eq'| `DOUBLE-Eq'| `STRING-Eq'| `INT-Range'| `DOUBLE-Range'| `STRING-Range'| `INT-Trend'| `DOUBLE-Trend' | `STRING-Trend'
	 <partitions> ::= <partitions> <exp> | EMPTY
	 <exp> ::= <exp> $\land$ <exp> | <exp> $\lor$ <exp> | 
	 <var name> <num-op> NUMBER | <var name> <string-op> STRING
	 <num-op> ::= == | != | > | < | <= | >=  
	 <string-op> ::= == | !=   
	 <threshold> ::=  NUMBER | EMPTY
	 <window> ::=   NUMBER | EMPTY
	 \end{lstlisting}
\end{footnotesize}

%Upon running the inference engine, two inputs are required: a plain-text variable definition file that is an implementation of the variable definition language and a continuous .csv-format trace file. 
%A third optional input is the priors file, which assigns a prior probability to the variables defined in the variable definition file. Without the priors file, the trace file will be preprocessed for all values available to a variable, and all values within each variable will be assigned a uniform probability. In our two sample systems, priors are gleaned from all traces using a frequency count.

$Equality$ predicates are of the form $var$ $EqOP$ $const$, where variable type can be   $int|float|string|bool$,
and \textit{EqOP: == $\mid$ !=}. This type of predicate encompasses earlier examples such as \textit{TargetDetected=TRUE}, \textit{Speed=0} and \textit{droneMode=Hovering}. 
For non-integer $Equality$ variables, we support fuzzy predicates with the addition  of a 
threshold such that $variable$ $EqOP$ $const \pm \delta$, such as $Acceleration = 9.8 \pm 0.01$. Additionally, we support disjoint intervals over a single variable through disjunction of predicates, such as \textit{droneMode}=\textit{Hovering} $\lor$ \textit{droneMode}=\textit{Translating}. 
This kind of predicate is effective at capturing explicit conditional states such 
as those embedded in a state machine.

$Range$ predicates are of the form $var$ $OP$ $const$, with  $OP:$ $\mid < \mid > \mid >= \mid <= $, and include conjunctions and disjunctions. 
$Range$ predicates can capture implicit states encoded in variables values. 
For example, a variable  $Latency$ has values that can be partitioned indicating a fast response $Latency<10$, a medium 
response $Latency\geq 10\land Latency<20$, or a slow response $Latency\geq 20$, and the system behavior may be conditioned differently across those ranges.

$Trend$ predicates are different in that they involve  state sequences, 
which is particularly valuable to capture tendencies over time.
These predicates apply a function to a sequence of values within a 
configured number of timesteps, referred to  as a window. 
$Trend$ predicates take the form of $f(var[window])$ $OP$ $const$. 
The function $f$ can be a simply average over the window, but it can also be more complex.
For example, one $f$  we use in our implementation calculates the 
derivative of the best fit quadratic polynomial function of the values in this series.
Given a  $window$, the predicate checks whether 
the derivative of a variable has increased in that window: 
$dvar_w>0$, has decreased: $dvar_w<0$, or remains constant: $dvar_w=0$.  
As another example that we implemented in our study, a predicate on the variable $TrustHumanOnSystem$  
checks whether trust is increasing, decreasing, or remains unchanged. 
Similarly, for whether a car is in \emph{autonomous} or \emph{manual} mode, the $changeMode$ variable could encode the direction in which the mode changed.

%%There are two aspects of the predicate patterns worth further higlightling. 

The examples thus far present predicate expressions over single variables,  however, predicates may be composed through conjunctions 
to define more complex  models.  
%For example, we can condition \textit{MissionState=PossibleTargetDetected} upon \textit{x-velocityHigh}, or upon \textit{x-velocityHigh $\land$ MachineState=Hover}. 
Also, although the three built-in patterns shown are rich enough for our study, they do not preclude the use of others patterns specified by developers with the only requirement being that they are to be able to be evaluated on the target traces (as per the $eval$ function of Algorithm \ref{alg:process-trace}).
 
%\ms{Update grammar to include multiple-predicate givens and constraints. Explain compound predicates and constraints} 

%Each variable must be defined in terms of how their values can be interpreted. An integer can be interpreted as a string if it is defined as such in the variable definition grammar. For example, a delta variable must be interpreted in terms of the rate of change from a collection of their values stemming from a given timestep, or a bound variable must be considered in terms of the range of values it fits into.  

%Because variable values are discrete, a prior probability mass function (pmf) must be described for each variable. If a %uniform pmf is not assumed, a file must be provided describing that variable's pmf which includes each value or bounded region that could be assumed by the variable.

%Additionally, a continuously-sampled trace with a uniform timestep between records must be provided to the engine. Non-continuous variables, such as user commands or warning messages, may be filled in with "nop" placeholders.

\section{STUDY}
Our study is meant to answer two questions:

%\se{@Meriel: re-read questions}

\begin{itemize}
	\item RQ1: \emph{What is the value-added of the generated conditional probabilistic invariants as compared to unconditioned invariants?} 
	We judge value by interpreting the top generated invariants  with and without conditional probabilities, and by comparing them against those generated by Daikon\footnote{We note that Daikon  does not consider prior probabilities, assumes certain thresholds and data distributions, and most important it only supports single-predicate joint probabilities (Daikon's mechanism to identify state partitions  is called ``conditional'' splitting but it is really computing a ``joint'' probability between a variable holding a value and an invariant).}.
	\item RQ2: \emph{What is the value-added of the generated conditional probabilistic invariants as additional data becomes available over time?}
	We explore how the conditional invariants change as additional data is gathered and more information becomes available, and judge their value in terms of their  surprise ratio. 
\end{itemize}

In the following sections, we first describe the setup of engine and scenarios, then address these two research questions.\footnote{A third research question regarding cost is addressed in the appendix.}

\subsection{Study Setup}

\begin{table*}
	\caption{Variables and Predicates in the Drone Study.}
	\vspace{-0.07in}
 	\begin{footnotesize}
	\label{Drone-Variable}
	\begin{tabular}{llp{12cm}}
		\hline
		Variable	& Predicate Type & Meaning
		\\ \hline \hline
		UserCommand &  String Equality & Command chosen by user from a predefined list.  Values: Default (None), Hover, KeepSweeping, LookCloser, ReqAuto, ReqManual, ReturnHome, Land \\ \hline
		MissionState&    String Equality &  Set of states describing mission status.   Values: Complete, InProgress, InsideSweepArea, OutsideSweepArea, PossibleTargetDetected, Suspended \\ \hline
		MachineState&  String Equality &  Set of states describing drone status.  Values of: FinishedBehavior, Hovering, Landing, LosingVicon, Manual, OutsideSweepArea, PossibleTargetDetected, Sweeping \\ \hline
		x-velocity	& Float Range &  Magnitude of the x-velocity in m/s. Range: 0-1.0 \\ \hline
		y-velocity	& Float Range &	Magnitude of the x-velocity in m/s. Range: 0-1.0 \\ \hline
		reaction$\_$time	& Float Range & Time in seconds user took to give a command after being prompted. Range: 0.0-14.0 \\ \hline
		sensor.status & Integer Equality &   Values:. 1=no target detected, 2=sensor ready to detect, 3=full target detected, 4=partial target detected \\ \hline
	\end{tabular}
	\end{footnotesize}
	\vspace{-0.07in}
\end{table*}

\begin{table*}
	\begin{footnotesize}
	\caption{Variables and Predicates in the Autonomous Car Study.}
	\vspace{-0.07in}
 	\label{Driving-Variable}
\begin{tabular}{llp{12cm}}
\hline
Variable     &Predicate Type    & Meaning                                                                                                                 \\ \hline \hline
Mode         & String Equality   & Driving mode that the simulated car is in. Values: Autonomous, Manual \\ \hline
WheelChange&Float Trend& Rate of change of wheel angle of the simulated car per second in degrees. Range: -360$-$360\\ \hline
Throttle &Float Range& Throttle applied to the simulated car in percentage. Range: 0-100 \\ \hline
Brake    &Float Range& Braking pressure applied to the  car in percentage. Range: 0$-$100  \\ \hline
Speed & Float & Velocity of the  car per second in $m/s^2$. Range: -5$-$5  \\ \hline
Event      &String Equality      & Event detected by the sensor of the  car. \\ & & Values: Pedestrian, Obstacle, Truck, Cyclist, False Alarm, None (nothing is detected) \\ \hline
TrustChange& Integer Trend& Change of trust level towards the autonomous driving system. Values: -5, -4, -3, -2, -1, 0, 1, 2, 3, 4, 5 \\ \hline
\end{tabular}
		\end{footnotesize}
\end{table*}

%\lu{This section has lots of details, but I get lost when reading it. What are the key messages we want to convey? I feel like using 1.5 pages to describe the study setup is too much...}

To answer the research questions, we required systems that met three requirements.
(1) They had to exhibit stochastic and conditional behavior, (2) they had to be amenable to the proposed analysis in that they generated a trace and were accessible enough for us to interpret the findings, and (3) they had to cover different sources of uncertainty 
and type of systems to help us understand whether the results would generalize.
We identified two systems and contexts that meet those criteria and are described in more detail in the next subsections:
(1) a drone performing a reconnaissance mission, and (2) an autonomous driving car interacting with a human driver.
Both systems and their executing scenarios were developed at [anonymized], they cover two distinct domains with the ground system uncertainty caused primarily by the drivers and sensors, and the aerial system caused by the sensors.  

We applied our inference approach to both systems, but to illustrate the results in the space available, we answer the first research question using just the invariants for the drone and second using the invariants for the autonomous vehicle.

%\lu{what is aerial and ground systems? These terminologies were not used before, please use drones and autonomous cars instead, to be consistent with before. And mention the drone first to keep the order consistent.}

%For each system we prepared configuration files and converted their system traces 
%into a standard format processable by the inference engine. As a general strategy, when building 
%the configuration files we favored including all potentially interesting variables even when we could not clearly anticipate their relation to other variables. 
%We were more conservative in defining ranges, favoring fewer ranges because smaller partitions of variable values were difficult to define meaningfully without significant empirical tuning. For example, the values of $reaction\_time$ were split into two ranges, $0 \leq reaction\_time < 7$ being fast and $reaction\_time \ge 7$ being slow. These ranges were determined through Jenks natural breaks optimization \cite{Jensk1967IYC} with slight adjustments to the resulting breakpoints to create uniform intervals.
%Ranked invariants were evaluated by surprise ratio, calculated by comparison of the posterior over prior. 

%Priors were calculated according to their frequency count in a disjoint subset of the traces for each system corresponding to limited contexts.  

\subsubsection{Drone ISR} 
The drone scenario is designed to mimic a simple intelligence/surveillance/reconnaissance (ISR) mission. 
The drone is a DJI Tello \cite{tello}    interfaced with a series of controllers built on Robot Operating System \cite{ros}.
The drone navigates in a controlled indoor flying cage equipped with a Vicon localization system \cite{vicon}.
The ISR mission starts with  the drone autonomously taking off at  randomly defined ``base'' 
coordinates and approaching a predefined sweep area. 
Once the sweep area is reached, the drone sweeps  for a QR code-marked ``target''.
The waypoint  navigation and sweep speed are adjusted by PID controllers. 
When a target is detected, the drone stops and queries the operator to confirm the target, which may require closer manual exploration. 
When a target is positively adjudicated, the drone returns to base. 
If the drone loses localization services, it hovers and queries the user for a decision. 
The user may either request manual control  to guide the drone back into the last localized position or request an emergency landing. If the user does not respond within a set time, the drone emergency lands. The targets were positioned outside of the localization area so, upon close examination of a target, the drone may lose localization services. 

\textbf{Traces.} A total of 34 runs containing 109 unique variables were collected using one operator and a variety of user inputs. Of those runs, 30 detected a possible target, 23 had the operator taking manual control, 20 had the operator issuing a closer examination command, 21 saw at least a partial loss of localization services, 20 saw a total loss of localization services, and 3 ended in emergency landings. 

%\se{@Meriel: talk to about this: What would it take to do the same study by select a subset of the runs that correspond to a particular deployment scenario? For example, we pick the runs without loss of control? This will directly address comments from 2 of the reviewers} \ms{Added the following paragraph: }
%This categorization of traces presents the opportunity to subdivide invariant generation according to the type of trace. This would necessitate an adjustment to the priors, which are now distributions from a subset of the traces instead of all traces, and an understanding that these invariants are implicitly conditioned upon the run type. Categorization of the run type could be programmatic or left to developer discretion.

Traces were captured using ROS bagging \cite{rosbag}, which produces a trace of timestamped variable-value pairs. 
Our data wrangling scripts  transform those bags into traces with variable-value pairs at every timestep using interpolation in .csv format.
Traces were on average 93.5s long per run.  
For the Drone study we only apply the inference engine once to illustrate the types of invariants that can be generated.

\textbf{Prior.} To mimic a development context in which traces are collected during in-house testing to generate a prior, a random subset of half of the available traces were selected. This set was disjoint from traces used to generate invariants. 
%17 of these drone runs were randomly selected and their traces were used to compute the priors.  

%Code coverage, in that all drone states should be represented in the aggregate of those traces, was a concern in that typical system verification test runs would likely involve prioritize code coverage. However, the majority of tests were run as "standard operating" traces, in that the drone was not forced to reach a certain state.

%\se{Is it true that we explored only 30? It seems to me that we would have explored many more now that we can have a conjunction of predicates in the givens. Please re-recheck these numbers.}

\textbf{Space of invariants.} The system developer, a co-author of this paper, used her system knowledge  to instantiate the three basic patterns provided by the specification language to define an initial set of predicates, and later enhanced the predicate specification to include some of the invariants produced by Daikon. In the end, a total of 30 unique predicates were specified. Given that specification, the inference engine combinatorially explored those predicates as part of a $Outcome$ and $Given$, while considering the constraints of the specification.  
%For example, if a predicate is marked as a Given, it may only be combined with other Given predicates to form the conditional.
%However, the developer chose to enable all predicates to be used as either Givens or Outcomes.
Starting with those predicates, the engine
 %These were then narrowed down to the 15 most potentially valuable, from which the top ten invariants were produced. 
generated 72,347 invariants. To illustrate this space,
Table \ref{Drone-Variable} lists the variables and predicates that are part of the ten invariants that will be reported in the next section.

\begin{figure}
	\centering
	\includegraphics[width=.5\columnwidth]{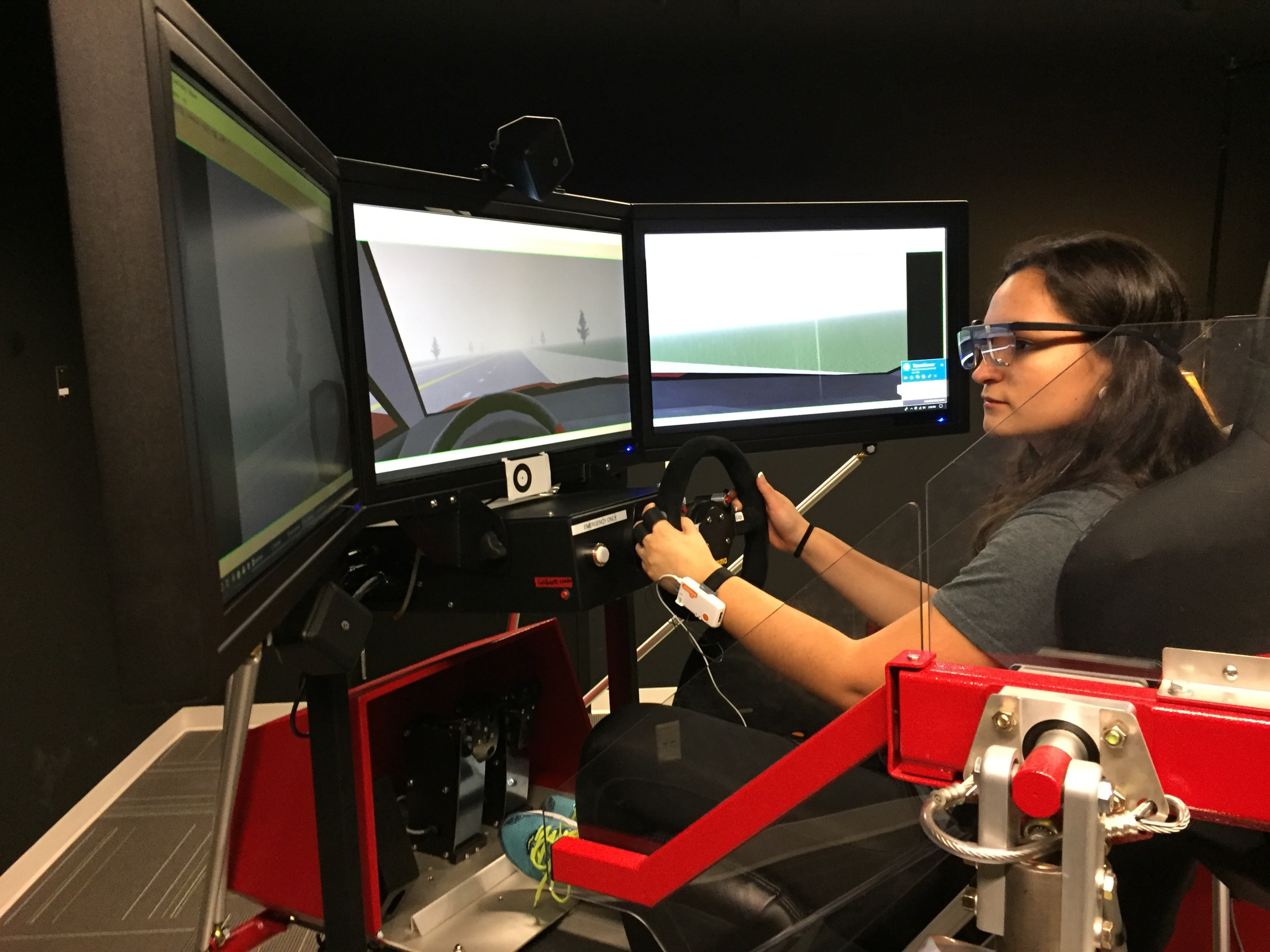}
	\caption{Autonomous driving scenario.}
	\label{fig:driving-scenario}
	\vspace{-0.2in}
\end{figure}

\subsubsection{Autonomous Driving}  \label{sec:auto-driving-study}
%\ms{Shili's section:  Setup, runs, number of trials, how drivers were chosen, what data was collected, what variables were considered interesting, etc.}
%\se{@Shili: please spell check / grammar check your text so that we do not have to do that}
%\se{Adressed@Shili: integrate both subsections into one, and make sure that every term in the tables is described in this section}
%\ms{Adressed@Shili: might want to mention how the Wheel, Brake, and Throttle values work in autonomous mode.}
\noindent The autonomous driving study is designed to explore drivers reactions  under 16 different scenarios, 
on a four-lane road where the participants  interact with a driving simulator \cite{forcedynamics}.
In each scenario, there are four potentially hazardous randomized
incidents: a pedestrian crossing the road, a cyclist riding slowly in the same lane, a stopped truck in the same lane, and an oncoming truck in the other lane. 
The simulated car is equipped with a sensor to detect events in the roadway within 40 meters and, depending on the scenario, the car may send an auditory alarm to alert the driver. 
%In four scenarios the system will send the auditory alarm to the driver before passing each incident. 
%In eight scenarios the system will send a false alarm (alarm activated while there is no incident). 
%In the remaining four scenarios the system will remain silent before any incidents.
Of the 16 driving scenarios, eight are fully-autonomous.
The other eight are semi-autonomous so the driver can switch between autonomous and manual driving mode. 
%The system will not respond to any operation from the human driver.% In the manual driving mode, the human driver operates the vehicle manually through the physical wheel, throttle, and brake.
The subjects can adjust their trust level towards the system from one (lowest) to five (highest).  
%The drivers can only change the trust level in the autonomous driving mode. 
Switching to manual driving mode will set trust level at 0. Switching from the manual to autonomous mode will set trust to three.

\textbf{Traces.} The study had 19 participants from [Anonymized]. %\footnote{[Anonymized]}. 
%The average age was 22.57 ($SD=$3.76). The gender ratio was 0.63 female. 
%All participants held a valid driver's license with more than one year's driving experience. 
Each participant had one training trial and 16 experimental trials.  9 participants were randomly selected and their driving traces (144 traces total) were used to compute the priors.
Traces were captured using the  PreScan software \cite{prescan} which was integrated with the simulator. 
Each trial lasted 180 seconds resulting in a trace of 180 elements with a time step of 1 second.
We recorded the vehicle dynamics, environment information, and user reactions through 32 variables. 

\textbf{Prior.} To mimic a development context in which traces are periodically collected from the field to revisit generated invariants, each trace was analyzed once at a time and then used as a source to refine the prior estimate.

\textbf{Space of invariants.} Experts in the system, some of which co-authored the paper, quickly fitted 7 variables of interest into the built-in predicate patterns, expressing also the combinations that seem less promising through the constraints. For example, because the occurrence of potentially hazardous incidents was predetermined in the scenarios, those variables could be limited to the givens as no action on the part of the driver or car could trigger them.
Based on the defined space, the inference tool generated 3,326 invariants.

\subsection{Results on Value-added for Conditional Invariants} \label{results}

We present the top ten invariants generated for the drone scenario in Table \ref{drone-invariant-table} ranked in descending order by surprise ratio.
% and top twenty invariants for the driving scenario in Figure \ref{fig:variance-of-surprise} 
(the rest of the invariants are available in the repo).
Invariants are presented in the form $P( Outcome \mid Given )$. The prior probability is the probability of $Outcome$ as calculated from the prior dataset (in this study that is computed only once as we only executed the generation algorithm once). The surprise ratio shows the relation of posterior to prior, and the explanation is a straightforward description of the invariant defined formally in the first column.

%\noindent \textbf{Drone ISR.} 
%\se{The presentation flow here is really hard to follow (tables are far from text describing them and each other, and we are asking the reader to follow us in multiple directions with little guidance that I even I had trouble figuring out what and how things were counted). }
%\se{A better flow to the presentation: Start by just introducing Table 7 first and describing two or three interesting invariants there, showcasing the surprise ratio, and how things that were hidden before now are visible. Then, introduce what is currently in Table 4: Priors,  discussing how if we were to focus on invariants with just those predicates we would miss the ones just discussed. Then, move to introduce the Daikon results and do a similar comparison than with Table 4.}
%Table \ref{drone-invariant-table} shows the top ten invariants for the drone scenario, 

%\se{@meriel: read carefully, section changed some parts}

\textbf{Top Invariants.} We start by interpreting the invariants reported in 
Table \ref{drone-invariant-table}. The system developer   expected   \textit{P(MissionState=Complete $\mid$ MachineState=Landing, UserCommand=Default,WarningState=Default)} on row 10. 
This is because in order for the drone to have completed its mission, it must make it back to its original starting point without incident and land. 
This invariant is strongly upheld by design, and so its probability will remain high across any traces supplied to the engine, especially when compared to the respective $0.05$ prior probability of $MissionState=Complete$.
%, which was calculated by frequency count across all state spaces in the prior dataset.

\begin{table*}[ht!]
	\begin{footnotesize}
		\caption{Drone Invariants}\label{drone-invariant-table}
		\vspace{-0.07in}
 		\begin{tabular}{l|l|l|l|p{2.8in}}

			\hline
			 \textbf{Invariant} & \textbf{Posterior} & \begin{tabular}[c]{@{}l@{}}\textbf{Original} \\ \textbf{ prior}\end{tabular} & \begin{tabular}[c]{@{}l@{}} \textbf{Surprise} \\ \textbf{ratio}\end{tabular}  & \textbf{Explanation} \\ \hline \hline
			\begin{tabular}[c]{@{}l@{}}P(sensor.status=4 $\mid$ \\ MissionState=PossibleTargetDetected \\ 0.01$\leq$y-velocity$<$0.25 reaction\_time=null )
			\end{tabular} &	0.30 & 0.002	& 159.85 &   When a possible target has been detected, y-velocity is low, and  no user reaction has been recorded, the sensor is likely detecting a partial target. \\ 
			\hline
			
			\begin{tabular}[c]{@{}l@{}}P(UserCommand=ReturnHome $\mid$ \\ MachineState=PossibleTargetDetected \\ x-velocity$\geq$0.25 )\end{tabular}&	0.52&	0.009	& 57.54 &  When a possible target has been detected and x-velocity is high, user has likely just issued a command to return home. \\  \hline
			
			\begin{tabular}[c]{@{}l@{}}P(UserCommand=Hover $\mid$ \\MachineState=Sweeping x-velocity$\geq$0.25 )\end{tabular} &	0.03 &	0.001 &	49.02 &   When drone is performing a sweeping task and x-velocity is high, user has likely just issued a command to hover. \\ \hline
			
			\begin{tabular}[c]{@{}l@{}}P(UserCommand=RequestAutoControl $\mid$ \\ MissionState=InProgress \\MachineState=LosingVicon \\WarningState=LosingVicon )\end{tabular}	& 0.28 &	0.01 &	45.93 &   When mission is in progress, drone has detected unreliable Vicon  connectivity, and a warning has been raised for unreliable Vicon  connectivity, the user has likely just issued a command to give  autonomous control to the drone.  \\ \hline
			
			\begin{tabular}[c]{@{}l@{}}P(UserCommand=Land $\mid$ \\ MachineState=Landing x-velocity$<$0.01 )\end{tabular} &	0.03	& 0.001	& 24.59 &  When machine is landing and x-velocity is low, the user has likely just issued a command to land.  \\ \hline
			
			\begin{tabular}[c]{@{}l@{}}P(WarningState=LosingVicon $\mid$ sensor.status=1 \\ MachineState=LosingVicon x-velocity$<$0.01 )\end{tabular} &	0.38	& 0.02 &	20.70 &   When sensor has not detected any target, drone has detected unreliable Vicon connectivity, and x-velocity is low,  a warning has   likely been raised for unreliable Vicon connectivity. 
			\\ \hline
			
			\begin{tabular}[c]{@{}l@{}}P(reaction\_time$>$7 $\mid$ \\ MissionState=OutsideSweepArea \\ sensor.status=3 y-velocity$\geq$0.25 )\end{tabular} &	0.23	& 0.01 &	19.85 &  When drone is outside the sweep area, the sensor has detected a  full target, and y-velocity is high, user reaction time is likely slow. \\
			\hline
			
			\begin{tabular}[c]{@{}l@{}}P(MachineState=FinishedBehavior $\mid$  \\ x-velocity$\geq$0.25 UserCommand=ReturnHome )\end{tabular} &	0.69	& 0.04 &	17.86 &   When x-velocity is high and user has issued a command to  return  home, the drone is likely finished its task.  \\ \hline
			
			\begin{tabular}[c]{@{}l@{}}P(MachineState=LosingVicon $\mid$ sensor.status=1 \\ x-velocity$<$0.01 WarningState=LosingVicon )\end{tabular} &	0.47	& 0.03 &	16.85 &   When sensor has not detected any target, x-velocity is low, and a  warning has been raised for no Vicon connectivity, the drone  has likely detected loss of Vicon connectivity. \\ 
			\hline
			
			\begin{tabular}[c]{@{}l@{}}P(MissionState=Complete $\mid$ MachineState=Landing \\ UserCommand=Default WarningState=Default )\end{tabular} &	0.76	& 0.05 &	16.72 &  When drone is landing, user has not issued a command, and no  warning has been raised, mission is likely complete.  \\ 
			\hline
			
			%\begin{tabular}[c]{@{}l@{}}P(WarningState=NoVicon | MissionState=Suspended \\ MachineState=Hovering UserCommand=Default ) \end{tabular} &	0.54425	& 0.03823	 &14.23626 & \begin{tabular}[c]{@{}l@{}} When mission is suspended, drone is hovering, and no user co-\\mmand has been issued, a warning has been raised for a loss of\\ Vicon connectivity.\end{tabular}\\ \hline
		\end{tabular}
		\vspace{-0.2in}
	\end{footnotesize}
\end{table*}

Some invariants had an unexpectedly high surprise ratio and were indeed surprising.
Invariant \textit{P(sensor.status=4 $\mid$ MissionState=PossibleTargetDetected,0.01$\leq$y-velocity$<$0.25, reaction\_time=null)} with the highest surprise ratio on row 1 showed that when the drone was in a \textit{PossibleTargetDetected} state and y-velocity was low, the sensor was only able to detect a partial target, 
%, no $UserCommand$ had been issued, and
perhaps indicating the need for a lower sweep speed in order to keep the full target in sensor range.
The high surprise ratio of \textit{P(UserCommand=RequestAutoControl $\mid$ MissionState=InProgress, MachineState=LosingVicon, WarningState=LosingVicon )} on row 4 was unexpected because other mission states can be associated with \textit{UserCommand=RequestAutoControl}, such as use of manual control to return to Vicon connectivity, which can occur during any machine state. This invariant tells us that the user had difficulty perceiving when localization connectivity was re-established and tried using autonomous control when it was not possible, which could be an opportunity for system improvement. \textit{P(reaction$\_$time$>$7 $\mid$ MissionState=OutsideSweepArea, sensor.status=3, y-velocity$\geq$0.25 )} on row 7 was also unexpectedly high. 
The predicate \textit{reaction$\_$time$>$7} can be associated with any command, but it seems to be most closely associated with end-of-mission commands as it is most closely related to \emph{MissionState=OutsideSweepArea}. The slow reaction time when the sensor is detecting a full target and the drone is outside the sweep area 
could indicate a need for sensor stabilization or a higher x-velocity at that time.

\textbf{Priors Comparison.}
When we compare the common predicates between invariants in Table \ref{drone-invariant-table} and priors in Table \ref{tab:priors}, we also find  marked dissimilarities. For example, predicates $flight\_data.battery\_low$, $y\mhyphen velocity<0.01$, and $0.01\leq x\mhyphen velocity < 0.25$ do not appear in the top ten conditional invariants.
%No predicates appear in both the top ten priors and top ten invariants' posterior outcome predicates. %, although the posterior givens predicates share 5 predicates with priors and 2 predicates with outcomes. 
Out of the 25 total unique predicates in Table \ref{drone-invariant-table}  only 5 also appear in the top ten priors.
This shows that stateful conditioning has a significant impact upon the probability of predicates.

%This shows that various subspaces within the trace show markedly different probabilistic behaviors than the priors characterizing the entire event space of similar traces, and which the naive approach discussed in the introduction could not capture. The wide gap in prior and posterior probabilities underlines the need to capture behavior on a stateful basis to more accurately represent system behavior.

%, since 7 out of the top 10 prior predicates were present in the top invariants. 
%All prior predicate probabilities were greater than or equal to 0.02941, and 0.4424 on average. 

%\se{Missed transition here, this is perhaps fixed with the adjusted flow I suggested above. I think these invariants could be some of the ones you still discussed as per the flow above.}

%We note that existing most trace-dependent invariant generation techniques rely on program structures when these stateful behaviors can often span multiple subroutines and do not report probabilistic certainty whatsoever. 

%\se{Need transition into Daikon here and why we are providing this comparison.}
\textbf{Daikon Comparison.}
We then  compare our approach to the perennially popular inference engine Daikon, primarily aiming to highlight the differences and the potential complementary nature of the approaches.
Table \ref{drone-invariant-table-daikon} shows the output for a tweaked version of  Daikon that includes the invariant probabilities for single-predicate splitting and allows for expression of stateful behavior as a conjunction.   
The probabilities of the \textit{UserCommand} predicates in Table \ref{drone-invariant-table-daikon} and Table \ref{drone-invariant-table}, 
show that the Daikon invariants capture probabilities closer to unconditioned prior probabilities, but give no indication of the circumstances under which those commands occur.
In the \textit{warning$\_$state$\_$change} entry point, we see that the range of \textit{x-velocities} is similar to the Bayesian invariant \textit{P(WarningState=LosingVicon $\mid$ sensor.status=1, MachineState=LosingVicon, x-velocity$<$0.01)} on row 6. 
However, this is not a conditional invariant but rather a conjunction of \textit{sensor.status==1 $\land$ x-velocity $\geq$ -0.609} which holds with confidence $\geq$0.95 at this program point. Moreover, it is not possible to split this program point using more than one predicate at a time, thus overlooking most of the invariants in   Table \ref{drone-invariant-table}.
%. This would require the consumer of these invariants to cross-reference across many split program points, and prevent the consumer from exploring relations between more than two predicates at a time.
%\se{Need to describe the table a bit more, what it captures, any overlap with ours?}
While these invariants are informative, they do not provide the conditional probabilities of our approach. Yet, Daikon provides a much richer set of builtin predicates than the ones covered by our implementation so we see much potential in their integration.

Overall, \textbf{the generated conditional probabilistic invariants confirm defined-by-design properties and expose properties of the drone ISR system that were heretofore unknown or not obvious and not captured by existing approaches. }

%\begin{figure}
%	\centering
%	\includegraphics[width=1\columnwidth, scale=1.5]{venn-drone}
%	\caption{Drone predicates in top priors vs. in top posteriors.}
%	\Description{}
%	\label{fig:venn-drone}
%	%\vspace*{-4mm}
%\end{figure} 

%\begin{figure}
%	\centering
%	\includegraphics[width=1\columnwidth, scale=1.75]{venn-driving}
%	\caption{Driving predicates in top priors vs. in top posteriors. \ms{Replace with max/min graphs of increasing number of predicates in givens}}
%	\Description{}
%	\label{fig:venn-driving}
%	%\vspace*{-4mm}
%\end{figure}

\begin{table}[]
	\caption{Top Prior probabilities for Drone.}
%	\vspace{-0.1in}
%	\subfloat[Drone scenario]{	\label{drone-priors}
\centering
		\begin{footnotesize}
			\begin{tabular}{lll}
				\hline
				\textbf{Predicate} & \textbf{Prior} \\ \hline
				UserCommand=Default &  0.958 \\ \hline
				flight$\_$data.battery$\_$low=False & 0.950 \\ \hline
				WarningState=Default & 0.897 \\ \hline
				status.data=1 & 0.724 \\ \hline
				MachineState=Sweeping & 0.427 \\ \hline
				y-velocity $<$ 0.01 & 0.426 \\ \hline
				MissionState=InsideSweepArea & 0.405 \\ \hline
				0.01 $\leq$ x-velocity $<$ 0.25 & 0.390 \\ \hline
				x-velocity $<$ 0.01 & 0.376 \\ \hline
				0.01 $\leq$ y-velocity $<$ 0.25& 0.354 \\ \hline
			\end{tabular}
		\end{footnotesize}
		%}
		% \subfloat[Driving scenario]{	\label{driving-priors}
		% 	\begin{footnotesize}
		% 		\begin{tabular}{lll}
		% 			\hline
		% 			\textbf{Predicate}  & \textbf{Prior} \\ \hline
		% 			Brake=0 & 0.956 \\ \hline
		% 			Event=None & 0.918 \\ \hline
		% 			Throttle$>$0 & 0.902 \\ \hline
		% 			Mode=Autonomous & 0.852 \\ \hline
		% 			WheelChange$\geq$20 & 0.504 \\ \hline
		% 			WheelChange$<$20 & 0.496 \\ \hline
		% 			VelocityChange$>$0 & 0.466 \\ \hline
		% 			VelocityChange$<$0 & 0.197 \\ \hline
		% 			Mode=Manual & 0.148 \\ \hline
		% 			TrustChange$<$0 & 0.120  \\ \hline
		% 			%Throttle=0  & 0.098 \shili{11} \\ \hline
		% 		\end{tabular}
		% 	\end{footnotesize}
		% }		\vspace{-0.2in}
		\label{tab:priors}
	%\vspace{-0.1in}
	\end{table}

	\begin{table}[]
		\begin{footnotesize}
			\caption{Daikon Drone Invariants}\label{drone-invariant-table-daikon}
			\vspace{-0.1in}
			\begin{tabular}{l}
				\hline
				\textbf{Daikon Drone Invariants}                                                                                                                                                      \\ \hline \hline
				\begin{tabular}[c]{@{}l@{}}/flight\_data.battery\_low one of \{ ``False'' (90.83\%), ``True'' (9.17\%) \}\end{tabular} \\ \hline
				%		/command\_state.data one of \{ "CommandState.Auto" (42.05\%), "CommandState.Default" (49.46\%), "CommandState.Manual" (8.49\%) \}                                                                                                                                                                                                                                                                                                   \\ \hline
				%		\begin{tabular}[c]{@{}l@{}}/machine\_state.data one of \{ "MachineState.FinishedBehavior" (4.74\%), "MachineState.Hovering" (4.32\%), "MachineState.Landing" (3.59\%), \\ "MachineState.LosingVicon" (2.79\%), "MachineState.Manual" (11.07\%), "MachineState.OutsideSweepArea" (15.47\%), "MachineState.PossibleTargetDetected" (17.96\%), \\ "MachineState.Sweeping" (38.47\%), "MissionState.Suspended" (1.30\%) \}\end{tabular} \\ \hline
				\begin{tabular}[c]{@{}l@{}}MissionState one of \{ ``Complete'' (4.63\%), ``InProgress'' (9.14\%),\\ ``InsideSweepArea" (37.34\%), ``OutsideSweepArea'' (22.76\%), \\ ``PossibleTargetDetected'' (18.21\%),  ``Suspended''(7.21\%), \\ ``AbortingMission'' (0.73\%) \}\end{tabular}                                                                 \\ \hline
				\begin{tabular}[c]{@{}l@{}}UserCommand one of \{ ``None'' (1.00\%), "Hover" (0.55\%), \\``KeepSweeping'' (2.51\%), ``Land" (0.73\%), ``LookCloser" (58.75\%),\\ ``RequestAutoControl" (11.79\%),  ``RequestManualControl'' (13.85\%), \\``ReturnHome'' (10.83\%)\} \end{tabular}                                                                     \\ \hline
				reaction\_time $\geq$ 0.0  \\ \hline
				%-1.0 $\leq$ x-velocity $\leq$ 1.0 \\ \hline                                                                                                                                                                                                                                                             
				\begin{tabular}[c]{@{}l@{}}..warning\_state\_change():::ENTER;condition=``sensor.status  ==  1''\\
					y-velocity $\leq$ 0.23\\
					x-velocity $\geq$ -0.61\end{tabular}
				\\ \hline
			\end{tabular}
		\end{footnotesize}
		\vspace{-0.3in}
	\end{table}

\subsection{Results on Value-added with Data Updates}

%\ms{Biggest changes here: }\noindent \textbf{Autonomous Driving.} 
%\ms{@Shili -- Analyze driving invariants table. Why are these invariants valuable?}
%\ms{@Shili: Please start spell checking.}
%\ms{unfinished sentence; relate priors and invariants tables}
%\ms{@Shili: Compare top invariants to top priors. Analyze invariants, don't just repeat what's in table.  E.g. invariants that were suprising, invariants that could lead to improvements in system. Read what I wrote about the drone invariants if you're lost.}
%\se{Need restructuring of flow, similar to the one proposed Drone section}
%\se{I did not check all the counts for Drone and Car. Double-check please}
Similar to the drone scenario, some of the initially generated invariants with  the highest posterior likelihood confirm our understanding of how the autonomous vehicle operates while others render interesting surprises.
For example, the invariant \emph{P(TrustChange$<$0 $\mid$ \\ Brake==0 Mode=autonomous \\ Throttle$>$0 Event=None WheelChange$\geq$20 )} says that, when the brake is not engaged, mode is autonomous, throttle is engaged,  nothing is detected in the roadway, and wheel angle is changing quickly, a decrease in trust is likely to occur. This may indicate that the controller for the car's angular motion makes drivers uncomfortable.
%\se{For completeness, it would be good to at least provide one great example of a car invariant we generated and describe it here.}
Also similar are the noticeable differences between conditioned and non-conditioned probabilities as shown by the differences between posteriors and priors. For example, of the outcome predicates, $TrustChange>0$, $TrustChange==0$, and $Brake>0$ do not appear in the priors. Of the givens, $Throttle==0$, $TrustChange>0$, and all $Event$ predicates excepting $Event=None$ do not appear in the priors.
%The rest of the posterior predicates related to $Mode$ and $WheelChange$ appear at least once in the top invariants.
%$Mode=Manual$, $WheelChange>=20$, and $Brake==0$ appear universally in all sets.

\begin{figure}
	\centering
	\includegraphics[width=\columnwidth]{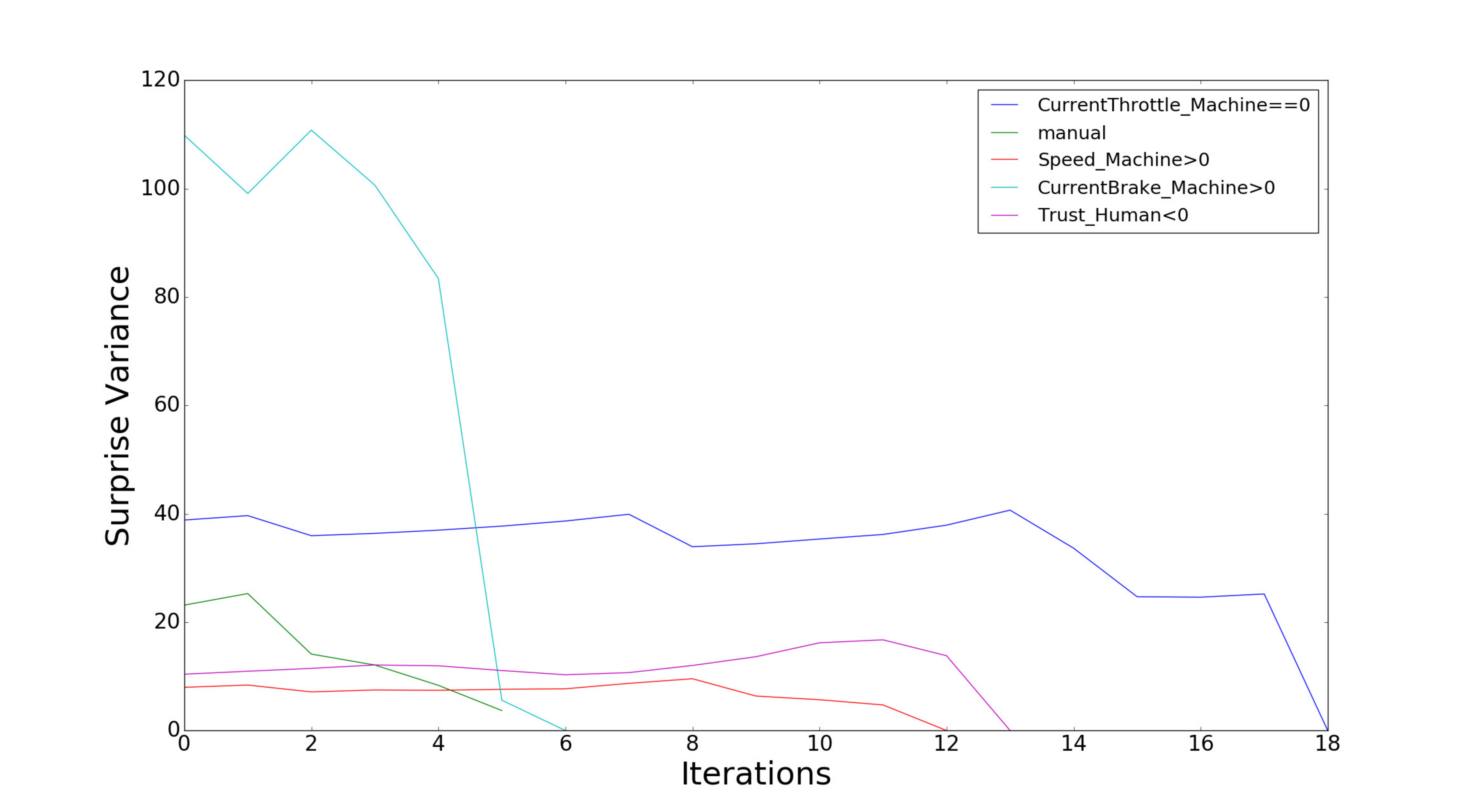}
	\caption{Variance of surprise ratio per iteration for driving scenario probabilistic invariants that exhibit high variance of surprise ratio.}
	\label{fig:variance-of-surprise}
	\vspace{-0.2in}
\end{figure}

However, in this question, we want to focus on the exploration of the effects of processing new traces and update priors in the generation and refinement of computed  invariants. 
%Figure \ref{fig:variance-of-surprise} summarizes the results for the top invariants selected for convergence of surprise ratios and low-entropy models. \se{We need to explain what those two terms means (not sure I get the criteria either) -- add a couple of sentences here}
%\ms{Added paragraph on Figure \ref{fig:variance-of-surprise}: }
To do that, we assume that the traces generated by each participant driving the autonomous vehicle are sequentially received and processed by the inference generation engine. So new traces are used to refine or create invariants   as well as to update priors for the next trace. Figure \ref{fig:variance-of-surprise} plots, for the 19 set of traces (invariant generation iterations in the x-axis), the variance of the surprise ratio of the generated invariants. 
%This figures 
%plots the accumulated variance of the invariant space's surprise ratios as the engine consumes new traces. 
A decrease in variance indicates that the surprise ratio of that particular invariant is converging, suggesting that the prior and posterior have both converged as well. 
%As we can see, most invariants' surprise variance drops off after the 6th or 7th iteration as the variability of user performance decreases. 
From there, model selection was used according to a modified Bayesian Information Criterion to constrain resulting invariants to low-entropy models. This was in order to determine whether the resulting distributions of prior and posterior were beginning to converge, hopefully affording some guarantee of predictability in future traces.

%We chose to use a convergence metric of a variance less than 0.01 of the accumulated average surprise to limit the number of invariants a developer might sift through. 
 
%\se{@Meriel: ADD we need a paragraph describing the figure in here. Something with this pattern: Some invariants like X show marked peaks that signify a big surprise because Y, some have a low variance throughout meaning ..., and some have a high variance earlier but converge over time as ...} 

Figure \ref{fig:variance-of-surprise} shows some invariants with marked peaks, indicating a large surprise potentially due to an aberrant trace, while some have low variance throughout, meaning that the posterior and prior do not change greatly between iterations. 

%\se{@Meriel: not sure if the following 3 paragraphs were meant as examples of what we see in the graph. If that is the case, they need to be linked (color line in graph corresponds to paragraph invariant or something along those lines). }

In the autonomous driving scenario, alarms are important to get drivers' attention and alert them to possible incidents. Invariant 
\textit{P(Brake$>$0 $\mid$ Mode=autonomous Event=pedestrian detected WheelChange$\geq$20 TrustChange$<$0 )}, 
the topmost light blue line in the graph, shows  a high degree of surprise variance early on in trace collection. This is mostly due to the fact that the initial prior is uniform, and in actuality pedestrians are an uncommon occurrence with a probability of $0.03$. This is the unconditioned probability by system design, and though there are three early traces in which this invariant occurs, the addition of more traces allows the prior to converge and thus the variance drops.

Trust affects human drivers' reliance on the system. \textit{P(Trust$<$0 $\mid$ Speed$>$0 Mode=autonomous Event=obstacle detected )} indicated by the purple line second closest to the x-axis, tells that when an obstacle is in the roadway, the car is under autonomous control, and the speed is nonzero, the trust level is more likely to decrease. The fact that the user must manually decrease the trust, either by self-reporting or by taking the car to manual mode, seems to indicate this is a perceived safety issue on the part of the human in the loop. Fully understanding the trust evolution contributes to a trustworthy system.

As with the previous drone study, the invariants generated for the autonomous vehicle confirmed expected known design attributes and helped to identify overlooked properties that characterize the system in the deployed situation.
Particularly relevant to this portion of the study, however, is how \textbf{new information can be easily processed by the  engine to generate new invariants, taking into consideration either new traces or updated prior distributions.} It is also worth noticing how, \textbf{at any given point in time, the surprise ratio is able to highlight different invariants as either the trace captures some new behavior or the prior changes, bringing to attention different aspects learned from the system.}

\section{CONCLUSIONS} 
\label{sec:conclusion}

In this work we have introduced what may be the first automated approach 
to generate conditional probabilistic invariants leveraging Bayesian inference. 
Our study showed its viability and potential in two distinct autonomous systems/contexts.
While most invariant use cases apply to probabilistic conditional invariants, 
we believe that a probabilistic understanding of stateful behavior can quantitatively reaffirm 
properties, assist in the discovery of unexpected behaviors appearing only under certain contexts, and expose potential opportunities to optimize a system with greater context specificity.

\section{APPENDIX}
%\documentclass[10pt,conference,anonymous]{IEEEtran} 
%\usepackage{comment}
%\usepackage{subfig} 
%\usepackage{graphicx}
%\usepackage{url}
%\usepackage{xcolor}
%\usepackage{multirow}
%\usepackage{listings}% http://ctan.org/pkg/listings
%\lstset{
%	basicstyle=\ttfamily,
%	mathescape,
%	columns=fullflexible
%}
% \usepackage{algorithmicx,algpseudocode}
%\usepackage{algpseudocode}
%\usepackage{array}
%\usepackage{enumitem}
%\usepackage[linesnumbered,algoruled,boxed,lined]{algorithm2e}
%\usepackage{amsmath}
%\mathchardef\mhyphen="2D % Define a "math hyphen"
%\usepackage{stmaryrd}
%\usepackage{cuted}
%\usepackage{syntax}
%\usepackage{multicol}
%
%
%
%\thickmuskip=0mu
%
%
%
%
%\begin{document}
% 
%\title{Probabilistic Conditional System Invariant Generation with Bayesian Inference: \\Appendix} 
%\author{Meriel Stein and Sebastian Elbaum}
%\authornote{Both authors contributed equally to this research.}
%\email{ms7nk@virginia.edu, selbaum@virginia.edu}
%\authornotemark[1]
%\affiliation{%
%  \institution{University of Virginia}
%  \city{Charlottesville}
%  \state{Virginia}
%  \postcode{22903}
%}
%
%\author{[Anonymized]\\Email\\Affiliation} 
%
% 
% 
%\maketitle

In this Appendix with provide additional study data and explanations that we could not fit in the paper submission.

\begin{table*}[h!t]
	\begin{footnotesize}
		\caption{Driving Invariants} \label{driving-invariant-table}
		 \begin{tabular}{l|l|l|l|p{2in}}
			\hline
			\textbf{Invariant} & \textbf{Posterior} & \textbf{Original prior} & \textbf{Surprise Ratio}    & \textbf{Explanation} \\ \hline \hline

			\begin{tabular}[c]{@{}l@{}} P(Brake$>$0 $\mid$ Mode=autonomous Throttle==0 \\ Event=pedestrian detected TrustChange$>$0 )\end{tabular} &	0.97	& 0.04	& 21.85 &   When mode is autonomous, throttle is not  engaged, a pedestrian is in  the roadway, and trust has increased, brake is likely engaged. \\ \hline
			
			\begin{tabular}[c]{@{}l@{}} P(Throttle==0 $\mid$ \\ Mode=autonomous Event=pedestrian detected ) \end{tabular} & 	1 &	0.1 	& 10.26 &   When mode is autonomous and a pedestrian is in the roadway, throttle  is likely not engaged.   \\ \hline
			
			\begin{tabular}[c]{@{}l@{}} P(Mode=manual $\mid$ \\ Throttle==0 Event=None )	 \end{tabular} & 0.95 &	0.15 &	6.40 &   When throttle is not engaged and nothing is detected in the roadway,  mode is  likely manual.  \\ \hline
			
			\begin{tabular}[c]{@{}l@{}} P(WheelChange$<$20 $\mid$ \\Mode=autonomous Event=None )	 \end{tabular} & 0.53 &	0.50 &	2.91 &   When mode is autonomous and nothing is detected in the roadway,  wheel angle is likely changing slowly. \\ \hline
			
			\begin{tabular}[c]{@{}l@{}} P(TrustChange$>$0  $\mid$ \\ Brake==0 Mode=manual Throttle$>$0 Event=None )	 \end{tabular} & 0.31 &	0.12 &	2.695 &   When brake is not engaged, mode is manual, throttle is engaged, and nothing is detected in the roadway, trust is likely increasing.  \\ \hline
			
			\begin{tabular}[c]{@{}l@{}} P(WheelChange$>=$20 $\mid$ \\ Brake==0 Mode=autonomous Throttle$>$0 \\ Event=cyclist detected TrustChange<0 ) \end{tabular} & 	0.90 	& 0.5 	& 1.79  &    When brake is not engaged, mode is autonomous, throttle is engaged, a cyclist is in the roadway, and trust increased, wheel angle is  likely  changing quickly.  \\ \hline
			
			\begin{tabular}[c]{@{}l@{}} P(TrustChange$<$0 $\mid$ \\ Brake==0 Mode=autonomous \\ Throttle$>$0 Event=None WheelChange$>=$20 ) \end{tabular} & 	0.15	& 0.12	& 1.21  &   When brake is not engaged, mode is autonomous, throttle is engaged,  nothing is detected in the roadway, and wheel angle is changing quickly, trust is likely to decrease. \\ \hline
			
			\begin{tabular}[c]{@{}l@{}} P(Throttle$>$0 $\mid$ \\Mode=autonomous Event=None ) \end{tabular} & 	1 &	0.90  &	1.11 &   When mode is autonomous and nothing is detected in the roadway,   throttle is likely engaged.  \\ \hline
			
			\begin{tabular}[c]{@{}l@{}} P(Brake==0 $\mid$ \\Mode=autonomous Event=None ) \end{tabular} & 	1 &	0.96	& 1.05 &   When mode is autonomous and nothing is detected in the roadway,  brake is likely not engaged.  \\ \hline
			
			\begin{tabular}[c]{@{}l@{}} P(TrustChange==0 $\mid$ \\Brake==0  Mode=autonomous Throttle>0 \\Event=None WheelChange$>=$20 ) \end{tabular} & 	0.77	& 0.77	& 1.01 &  When brake is not engaged, mode is autonomous, throttle is engaged,   nothing has been detected in the roadway, and wheel angle is changing  quickly, trust is likely to remain constant. \\ \hline
		\end{tabular}
	\end{footnotesize}
\end{table*}

\section{Autonomous Vehicle Invariants}

This analysis is relevant to the second study in the paper.

Table \ref{driving-invariant-table} includes the top 10 invariants generated for the autonomous vehicle. We briefly comment on them here. 

Invariant \textit{P(Brake$>$0 | Mode=autonomous Throttle==0 Event=pedestrian detected TrustChange$>$0 )} on row 1 had the highest surprise ratio, possibly due to the specific behavior the autonomous  controller exhibited in the presence of a pedestrian and the trust-building effect it had on the human in the loop. 

Invariant \textit{P(Throttle==0 $\mid$ Mode = autonomous Event=pedestrian detected )} on row 2 had a similarly high surprise ratio, with posterior likelihood of 1.0 indicating that when the car is autonomously controlled and a pedestrian is in the roadway, the throttle is no longer engaged. 

This has a slightly higher probability than \textit{P(Brake$>$0 $\mid$ Mode=autonomous Throttle==0 Event=pedestrian detected TrustChange $>$0 )}, likely because the throttle must be disengaged before brake can be engaged, and the brake may be engaged for multiple reasons, such as a different event or a curve in the road. The inclusion of predicate $TrustChange>0$ in the givens was surprising as well, as it indicates that an increase in trust in conjunction with detecting a pedestrian is correlated with a subsequent application of the brakes.
Note that this is not a causal relationship, as the autonomous driving algorithm does not react to changes in trust from the human in the loop. 

Dealing with incidents on the road is essential to demonstrate realistic safety behaviors in the autonomous driving scenario. Invariants given \textit{Event=Pedestrian detected} characterize the performance of the simulated car when handling the incident of pedestrian crossing the road. The car decreases the velocity by applying brake and easing the throttle, but no wheel change is closely associated according to the model selection algorithm. It fits the expectation that the car tends to slow down, rather than changing lanes to bypass the pedestrian. In larger models, both $WheelChange>=20$ and $WheelChange<20$ predicates present with no significant change to the posterior likelihood, showing that the posterior likelihood is more strongly conditioned on a change in $Throttle$. 

On the other hand, invariants given \textit{Event=Cyclist detected} shows that the car performs a steep turn and unexpectedly accelerates in order to avoid the cyclist. Outside of the top ten, similar invariants can be found involving $Event=Truck$, which shows that the car again unexpectedly accelerates when passing the incoming truck on the other lane. These invariants present trends that the designers were unaware of and provide direct guidance on improving the system design when handling incidents.  

In the autonomous driving scenario, alarms are important to alert drivers and get their attention to possible incidents. We intentionally injected false alarms to test the car's performance. The expected invariant $P(-20<=WheelChange<=20 \mid Event=False Alarm)$ tells that the car maintains slight change of the wheel, while the unexpected invariant $P(VelocityChange<0 \mid Event=False Alarm)$ indicates a velocity decrease. Recall that false alarm is an auditory alarm when there is no real incident on the road. Even though the false alarm does not cause the car to react too much on the wheel, it causes the car to slow down to check if there is any incident. False alarms improve the safety of the system by detecting potential hazards more conservatively, though too many false alarms may lead to driver's alarm fatigue.  

Invariants related to $Mode$ provide the information during manual driving or autonomous driving. Invariant \textit{P(TrustChange$>$0 $\mid$ Brake==0 Mode=manual Throttle$>$0 Event=None )} on row 5 shows that human drivers tend to increase their trust following ``normal'' operation in manual mode. This also indicates that they subsequently leave manual mode, as it is not possible to raise trust in manual mode. 

$P(Mode=Manual \mid Throttle=0)$ with a ratio of 3.986 and $P(Mode=Autonomous \mid Throttle=0)$ with a ratio of 0.481 tells that given throttle is not used, the car is more likely to be in manual driving mode. These invariants shows that human drivers tends to be unsatisfied by the speed of the autonomous driving and they are confident of driving manually. This can help the designer of the system to understand human driver's behavior and adjust the performance of the system.

Trust affects human drivers' reliance on the system. \textit{P(TrustChange $<$ 0 $\mid$ Brake==0 Mode=autonomous Throttle$>$0 Event=None WheelChange $>=$ 20 )} on row 7 tells that when throttle is engaged and the wheel angle is changing quickly, the trust level is more likely to decrease. The absence of an $Event$ seems to indicate this is a perceived safety issue on the part of the human in the loop. 

However, \textit{P(TrustChange==0 $\mid$ Brake==0 Mode=autonomous Throttle$>$0 Event=None WheelChange $>=$ 20 )} on row 10 shows us that the human in the loop is more likely to leave trust unchanged under those same conditions. This seems to indicate that there are latent variables not being accounted for, possibly in the system or on the part of the user. Fully understanding the trust evolution contributes to a trustworthy system.

 \section{Preliminary Results on  Inference Computation Cost}  \label{sec:results-cost}

We  explore \emph{What is the cost of generating conditional invariants?}.  
We assess cost in terms of the time to generate the invariants as a function of the number of predicates and trace length. 

 \begin{figure}
	\centering
	\includegraphics[width=.9\columnwidth, scale=0.25]{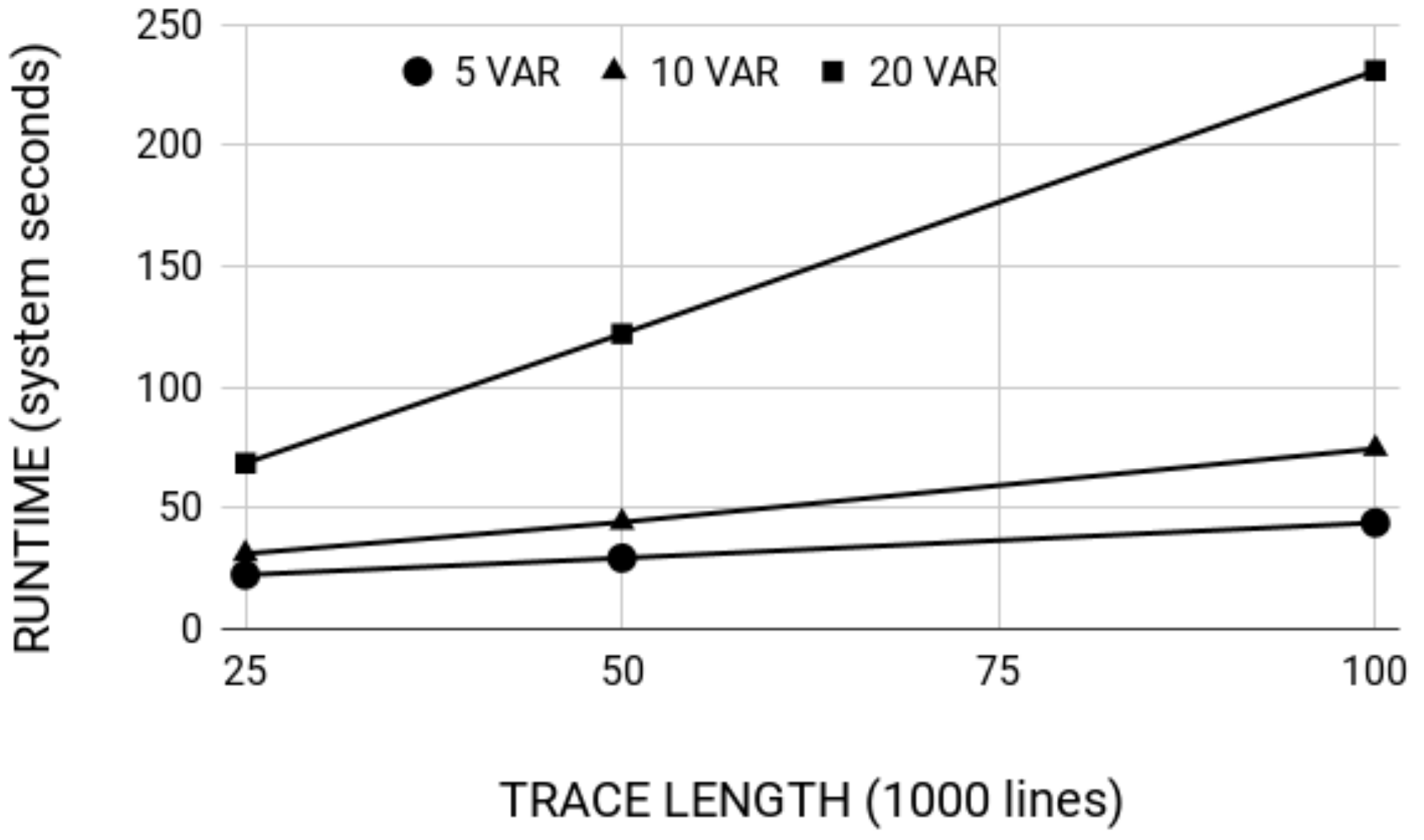}
	\caption{Runtime vs. trace length and number of predicates.}
	\label{fig:runtime-chart}
 \end{figure}  

 We briefly assess the  dimensions of the space of invariants to explore and the associated inference  cost. 
 Figure \ref{fig:runtime-chart} plots the runtime cost in system seconds
 when executing the inference engine on 
 traces of three different lengths (25K, 50K, 100K) produced by the Drone ISR when three different
 sets of predicates are explored (5, 10 and 20 variables of $Range$ type with two predicates each). 
 Runtime tests were performed on a containerized Linux box with an x86$\_$64 AMD FX-8120E 3.1GHz 8-core processor.
 
 As the graph shows, the runtime of the engine depends on both trace length and predicate complexity, but the influence of the number of variables seems  to dominate  the space to explore as it  grows exponentially when the variables are considered as both outcomes and givens.
 As the number of variables grows, a developer can control this cost through the specification of the predicate space provided to the engine by  
 specifying whether a variable is to be explored as a given or as an outcome,
 or more restrictively by aiming for particular pairs of variable predicates. 
 
 %In our analysis, the average of posterior probabilities across all traces was used to get a sense of the posterior within the trace population, as opposed to evaluating posteriors on a trace-by-trace basis. Using the average means that the averages of the values within a certain given condition may not add up to 1.00 due to those conditions being absent in some traces, causing the total probability to be zero. For example, $P(BatteryLow \mid NoVicon)$ and $P(BatteryOkay \mid NoVicon)$ may not add up to 1.00 if, in some traces, Vicon connectivity is never lost. 

%\end{document} 

\bibliographystyle{plain}
\bibliography{references}

\end{document}